\documentclass[review]{elsarticle}

\usepackage{lineno,hyperref}
\modulolinenumbers[5]

\usepackage{times}
\usepackage{epsfig}
\usepackage{graphicx}
\usepackage{subcaption}
\usepackage{amsmath}
\usepackage{amssymb}
\usepackage{hhline}
\usepackage{adjustbox}
\usepackage{multirow}
\biboptions{authoryear}

\journal{Neural Networks}

\begin{document}
\begin{frontmatter}
%
\title{\textit{Soft} + \textit{Hardwired} Attention: An LSTM Framework for Human Trajectory Prediction and Abnormal Event Detection}


\author[mymainaddress]{Tharindu Fernando \corref{mycorrespondingauthor}}
\cortext[mycorrespondingauthor]{Corresponding author}
\ead{t.warnakulasuriya@qut.edu.au.}

\author[mymainaddress]{Simon Denman}
\author[mymainaddress]{Sridha Sridharan}
\author[mymainaddress]{Clinton Fookes}

\address[mymainaddress]{Image and Video Research Laboratory, SAIVT, \\ Queensland University of Technology, \\Australia.}

\begin{abstract}
As humans we possess an intuitive ability for navigation which we master through years of practice; however existing approaches to model this trait for diverse tasks including monitoring pedestrian flow and detecting abnormal events have been limited by using a variety of hand-crafted features. Recent research in the area of deep-learning has demonstrated the power of learning features directly from the data; and related research in recurrent neural networks has shown exemplary results in sequence-to-sequence problems such as neural machine translation and neural image caption generation. Motivated by these approaches, we propose a novel method to predict the future motion of a pedestrian given a short history of their, and their neighbours, past behaviour. The novelty of the proposed method is the combined attention model which utilises both ``soft attention'' as well as ``hard-wired'' attention in order to map the trajectory information from the local neighbourhood to the future positions of the pedestrian of interest. We illustrate how a simple approximation of attention weights (i.e hard-wired) can be merged together with soft attention weights in order to make our model applicable for challenging real world scenarios with hundreds of neighbours. The navigational capability of the proposed method is tested on two challenging publicly available surveillance databases where our model outperforms the current-state-of-the-art methods. Additionally, we illustrate how the proposed architecture can be directly applied for the task of abnormal event detection without handcrafting the features.
\end{abstract}


\end{frontmatter}
\linenumbers

\section{Introduction}
Understanding and predicting crowd behaviour in complex real world scenarios has a vast number of applications, from designing intelligent security systems to deploying socially-aware robots. Despite significant interest from researchers in domains such as abnormal event detection, traffic flow estimation and behaviour prediction; accurately modelling and predicting crowd behaviour has remained a challenging problem due to its complex nature. \par
As humans we possess an intuitive ability for navigation which we master through years of practice; and as such these complex dynamics cannot be captured with only a handful of hand-crafted features. We believe that directly learning from the trajectories of pedestrians of interest (i.e. pedestrian who's trajectory we seek to predict) along with their neighbours holds the key to modelling the natural ability for navigation we posses. \par
The approach we present in this paper can be viewed as a data driven approach which learns the relationship between neighbouring trajectories in an unsupervised manner. Our approach is motivated by the recent success of deep learning approaches (\cite{GoroshinBTEL15, MadryBKF14, LaiBF14}) in unsupervised feature learning for classification and regression tasks. \par

\subsection{Problem Definition}
The problem we have addressed can be defined as follows:  Assume that each frame in our dataset is first preprocessed such that we have obtained the spatial coordinates of each pedestrian at every time frame. Therefore the trajectory of the ${i^{th}}$ pedestrian for the time period of ${1}$ to ${T_{obs}}$ can be defined as,    
\begin{equation}
 \mathbf{x_i}=[x_1, y_1, \ldots, x_{T_{obs}},y_{T_{obs}} ] .
 \label{eq:input_traj}
\end{equation}
The task we are interested in is predicting the trajectory of the ${i^{th}}$ pedestrian for the period of ${T_{obs+1}}$ to ${T_{pred}}$, having observed the trajectory of the ${i^{th}}$ pedestrian from time ${1}$ to ${T_{obs}}$ as well as the trajectories all the other pedestrians in the local neighbourhood during that period. This can be considered a sequence to sequence prediction problem where the input sequence captures contextual information corresponding to the spatial location of the pedestrian of interest and their neighbours, and the output sequence contains the predicted future path of the pedestrian of interest.

\subsection{Proposed Solution}
\begin{figure*}[t]
    \centering
     \includegraphics[width=.95\textwidth]{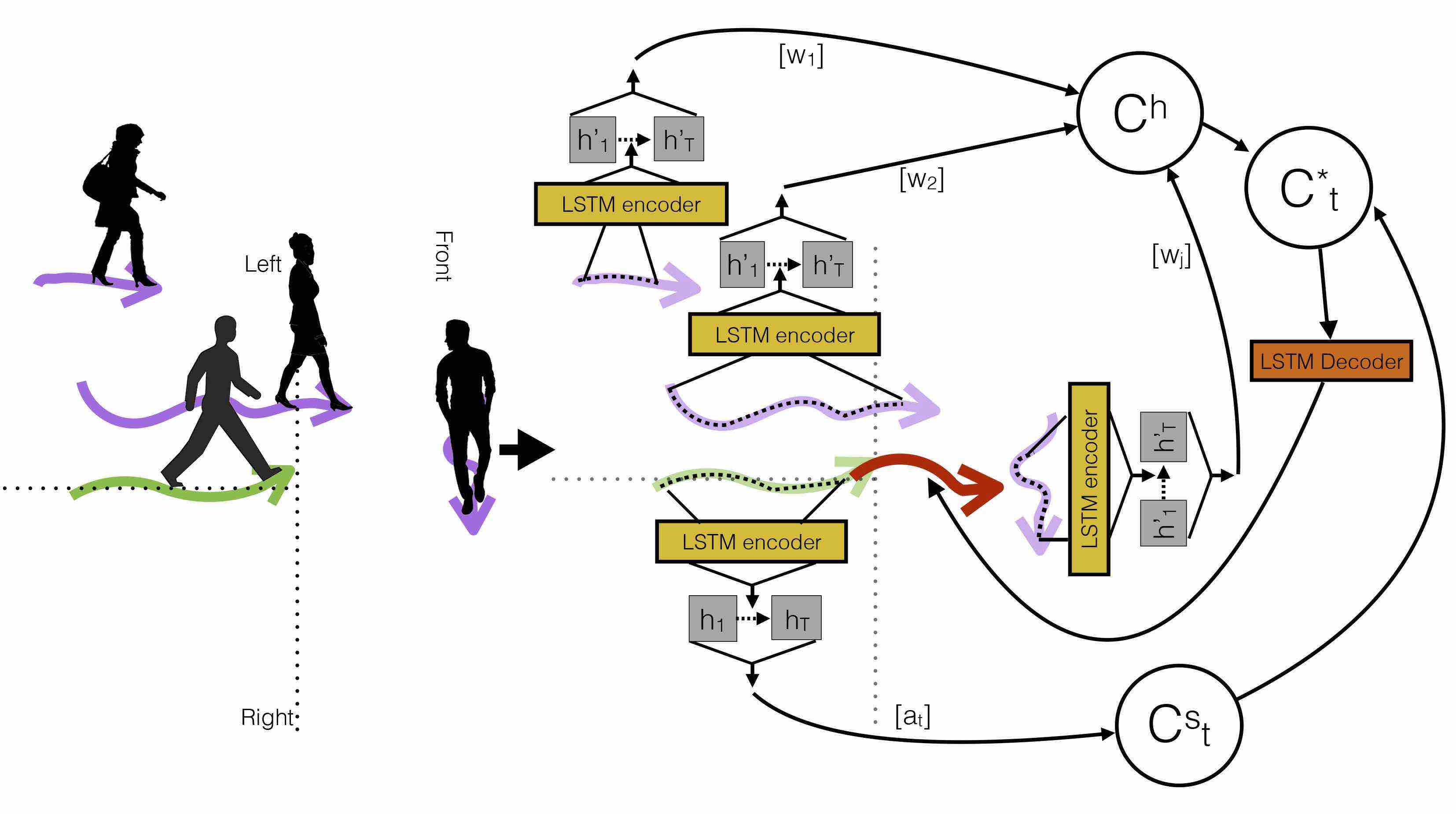}
			\vspace{-1mm}
        \caption{ A sample surveillance scene (on the left): The trajectory of the pedestrian of interest is shown in green, and has two neighbours (shown in purple) to the left, one in front and none on right. Neighbourhood encoding scheme (on the right): Trajectory information is encoded with LSTM encoders. A soft attention context vector $C^{s}_{t}$ is used to embed the trajectory information from the pedestrian of interest, and a hardwired attention context vector $C^{h}$ is used for neighbouring trajectories.  In order to generate $C^{s}_{t}$ we use a soft attention function denoted $a_t$ in the above figure, and the hardwired weights are denoted by $w$. The merged context vector is then used to predict the future trajectory for the pedestrian of interest (shown in red).}
        \label{fig:model_learning}
 \vspace{-3mm}
 \end{figure*}

To solve this problem we propose a novel architecture as illustrated in Fig. \ref{fig:model_learning}. 
For encoding and decoding purposes we utilise Long-Short Term Memory networks (LSTM) due to their recent success in sequence to sequence prediction \citep{NeuTrans, Xu_NIC_2015, YooPLPK15}.
 We demonstrate the social navigational capability of the proposed method on two challenging publicly available surveillance databases. We demonstrate that our approach is capable of learning the common patterns in human navigation behaviour, and achieves improved predictions for pedestrians paths over the current state-of-the-art methodologies. Furthermore, an application of the proposed method for abnormal human behaviour detection is shown in Section 5.
 
\section{Related Work}
\subsection{Trajectory Clustering}
When considering approaches for learning motion patterns through clustering, \citet{Giannotti_2007} have proposed the concept of ``trajectory patterns'', which represents the descriptions of frequent behaviours in terms of space and time. They have analysed GPS traces of a fleet of 273 trucks comprising a total of 112,203 points. 
Deviating from discovering common trajectories, \citet{Lee:2007} proposed to discover common sub-trajectories using a partition-and-group framework. The framework partitions each trajectory into a set of line segments, and forms clusters by grouping similar line segments based on density. \citet{Morris_2009} evaluated different similarity measures and clustering methodologies to uncover their strengths and weaknesses for trajectory clustering. With reference to their findings, the clustering method had little effect on the quality of the results achieved; however selecting the appropriate distance measures with respect to the properties of the trajectories in the dataset had great influence on final performance.\par
\subsection{Human Behaviour Prediction}
When predicting human behaviour the most common motion models are social force models \citep{Helbing_SF, KoppulaS13, PellegriniEG10, conf/cvpr/YamaguchiBOB11, Jingxin_2012} which generate attractive and repulsive forces between pedestrians. Several variants of such approaches exist. \citet{AlahiRF14} represents it as a social affinity feature by learning the pedestrian trajectories with relative positions where as \citet{YiLW15} observed the behaviour of stationary crowd groups in order to understand crowd behaviour.
With the aid of topic models the authors in \citet{WangMNG08} were able to learn motion patterns in crowd behaviour without tracking objects. This approach was extended to incorporate spatio-temporal dependencies in \citet{HospedalesGX09} and \citet{EmonetVO11}.\par
Deviating from the above approaches, a mixture model of dynamic pedestrian agents is presented by \citet{ZhouTW15}, who also consider the temporal ordering of the observations. Yet, this model ignores the interactions among agents, a key factor when predicting behaviour in real world scenarios. \par
The main drawback in all of the above methods is that they utilise hand-crafted features to model human behaviour and interactions. Hand-crafted features may only capture abstract level semantics of the environment and they are heavily dependent on the domain knowledge that we posses. 
\par
In \citet{social_LSTM} an unsupervised feature learning approach was proposed. The authors have generated multiple LSTMs for each pedestrian in the scene at that particular time frame. They have observed the position of all the pedestrians from time ${1}$ to ${T_{obs}}$ and predicted all of their positions for the period ${T_{obs+1}}$ to ${T_{pred}}$. 

They have pooled the hidden states of the immediately preceding time step for the neighbouring pedestrians when generating their positions in the current time step. A more detailed comparison of this model with our proposed model is presented in Sec. 3.3
\subsection{Attention Models}
Attention-based mechanisms are motivated by the notion that, instead of decoding based on the encoding of a single element or fixed-length part of the input sequence, one can attend a specific area (or important areas) of the whole input sequence to generate the next output. Importantly, we let the model learn what to attend to based on the input sequence and what it has produced so far. \par
In \citet{NeuTrans}  have shown that attention-based RNN models are useful for aligning input and output word sequences for neural machine translation. This was followed by the works by \citet{Xu_NIC_2015} and \citet{yao_VC_2015} for image and video captioning respectively. According to \citet{SharmaKS15}, attention based models can be broadly categorised into soft attention and hard attention models, based on the method that it uses to learn the attention weights. \textit{Soft attention} models \citep{NeuTrans, Xu_NIC_2015, yao_VC_2015, SharmaKS15} can be viewed as ``supervised'' guiding mechanisms which learn the alignment between input and output sequences through backpropagation. \textit{Hard attention} \citep{MnihHGK14,Williams92simplestatistical} is used by Reinforcement Learning to predict an approximate location to focus on. With reference to \citet{SharmaKS15}, learning hard attention models can become computationally expensive as it requires sampling.\par
Still, soft aligning multiple feature sequences is computationally inefficient as we need to calculate an attention value for each combination of input and output elements. This is feasible in cases such as neural machine translation where we have a 50-word input sequence and generate a 50-word output sequence, but prohibitively expensive in a surveillance setting when a target has hundreds of neighbours, and we have to learn the attention weight values for all possible value combinations for each of the neighbouring trajectories. We tackle this problem via the merging of soft attention and hardwired attention in our framework.

\section{Proposed Approach}
\subsection{LSTM Encoder-Decoder Framework}
In contrast to past attention models \citep{NeuTrans,Xu_NIC_2015} which align a single input sequence with an output sequence, we need to consider multiple feature sequences in the form of trajectory information from the pedestrian of interest and their neighbours when predicting the output sequence. Aligning all features together is not optimal as they have different degrees of influence (i.e. a person walking directly next to the target has greater influence than a person several meters away). Soft attention models are deterministic models which are trained using back-propagation. Therefore, aligning each input trajectory sequence separately via a separate soft attention model is computationally expensive. \par
We show that we can overcome this problem with a set of hardwired weights which we calculate based on the distance between each neighbour and the pedestrian of interest. When considering navigation, as distance is the key factor which determines the neighbour's influence, it acts as a good generalisation.  \par
\begin{figure*}[t]
    \centering
     \includegraphics[width=.95\textwidth]{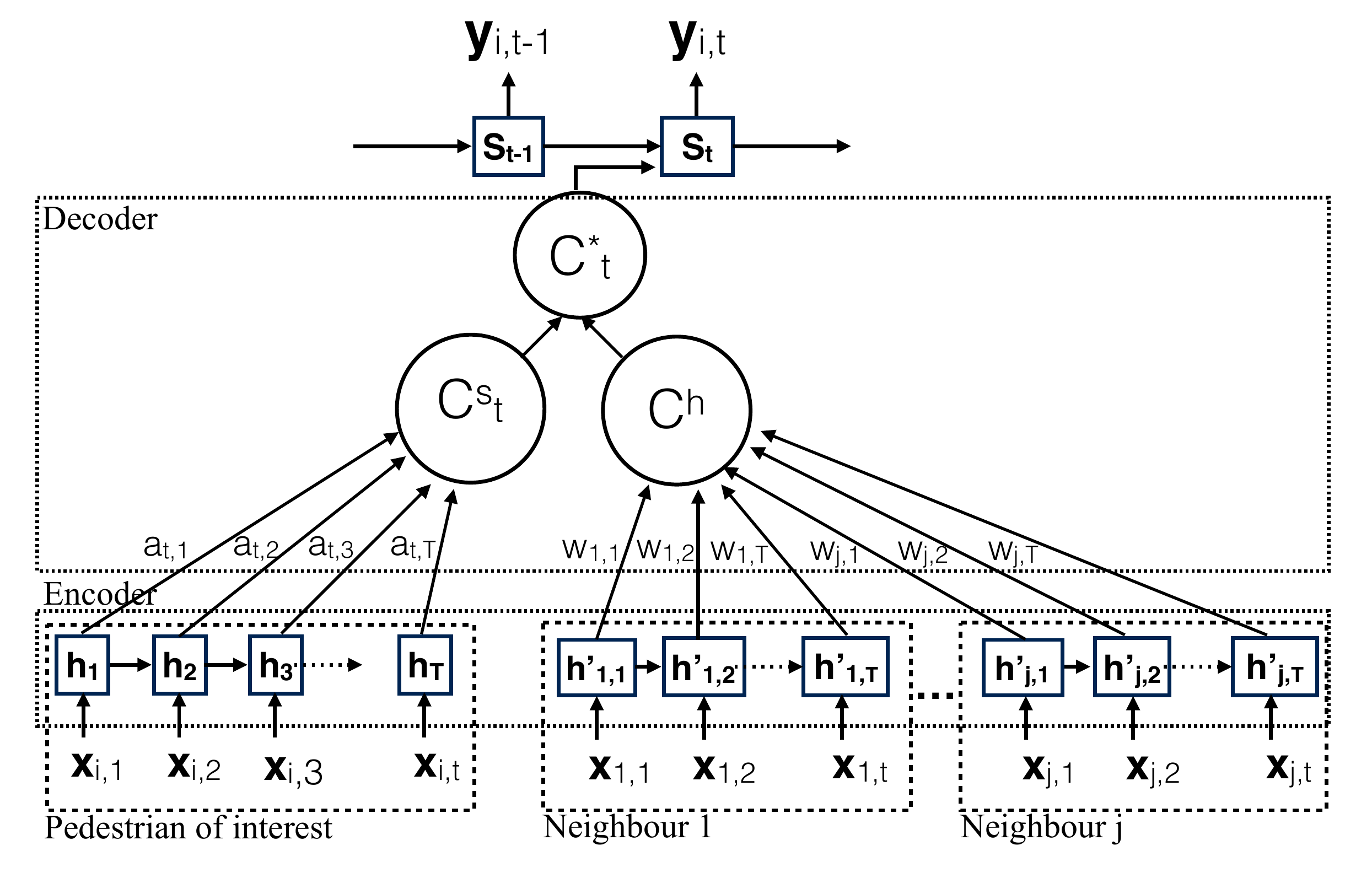}
        \caption{The proposed \textit{Soft} + \textit{Hardwired} Attention model. We utilise the trajectory information from both the pedestrian of interest and the neighbouring trajectories. We embed the trajectory information from the pedestrian of interest with the soft attention context vector $C^{s}_{t}$, while neighbouring trajectories are embedded with the aid of a hardwired attention context vector $C^{h}$. In order to generate $C^{s}_{t}$ we use a soft attention function denoted $a_t$ in the above figure, and the hardwired weights are denoted by $w$. Then the merged context vector, $C^{*}_{t}$, is used to predict the future state $\mathbf{y}_{i(t)}$}
        \label{fig:model}
 \end{figure*}
 \par 
 The proposed LSTM Encoder-Decoder framework is shown in Fig. \ref{fig:model}. Due to its computational complexities, soft attention (denoted $a_t$ in Fig. \ref{fig:model}) is used only when embedding the trajectory information from the pedestrian of interest. We show that by approximating the required attention for the neighbours through hardwired attention weights ($w_j$), which we calculate based on the distance between the neighbouring pedestrian and the pedestrian of interest, we can generate a good approximation of their influence. The methodology of generating soft and hardwired attentions is outlined in the following subsections.
\subsubsection{LSTM Encoder}
In a general Encoder-Decoder framework, an encoder recieves an input sequence $\mathbf{x}$ from which it generates an encoded sequence $h$. In the context of this paper, the input sequence for the pedestrian $i$ is given in Equation \ref{eq:input_traj} and the encoded sequence is given by,
\begin{equation}
 h_i=[h_1, \ldots, h_T ] .
\end{equation}
The encoding function is an LSTM, which can be denoted by,
\begin{equation}
h_{t}=\mathrm{LSTM}(\mathbf{x}_{t},h_{t-1}) .
\end{equation}
With the aid of above equation we encode the trajectory information from the pedestrian of interest as well as each trajectory in the local neighbourhood. 

\subsubsection{LSTM Decoder}
Before considering how the combine context vector $C^{*}_{t}$ is formulated, the concept of time dependent context vector can be illustrated as follows. For a general case, let $s_{t-1}$ be the decoder hidden state at time $t-1$, $\mathbf{y}_{t-1}$ be the decoder output at time $t-1$, $C_{t}$ be the context vector at time $t$ and $f$ be the decoding function. The decoder output at time $t$ is given by, 
\begin{equation}
\mathbf{y}_{t}=f(s_{t-1},\mathbf{y}_{t-1},C_{t}),
\label{eq:4}
\end{equation}
as defined by \citet{NeuTrans} such that distinct context vectors are given for each time instant. The context vector depends on the encoded input sequence $h=[h_1, \ldots, h_t]$.  \par

In the proposed approach, the given trajectory (i.e $\mathbf{x}=[x_1, y_1, \ldots, x_{T_{obs}},y_{T_{obs}} ]$) for the pedestrian of interest is encoded and used to generate a soft attention context vector, $C^{s}_{t}$. With the aid of distinct context vectors we are able to focus different degrees of attention towards different parts of the input sequence, when predicting the output sequence. The soft attention context vector $C^{s}_{t}$ can be computed as a weighted sum of hidden states, 
\begin{equation}
C^{s}_{t}=\sum_{j=1}^{T_{obs}} \alpha_{tj}h_j .
\end{equation}
In \cite{NeuTrans}, the authors have shown that the weight $\alpha_{tj}$ can be computed by,
\begin{equation}
\alpha_{tj}=\cfrac{exp(e_{tj})}{\sum_{k=1}^{T} exp(e_{tk})} , 
\end{equation}
\begin{equation}
e_{tj}=a(s_{t-1},h_j) ,
\end{equation}
and the function $a$ is a feed forward neural network for joint training with other components of the system.

As an extension to the decoder model proposed in \citet{NeuTrans}, we have added a set of hardwired weights which we use to generate \textit{hardwired attention} $ (C^{h})$. 
Utilising the hardwired attention model we combine the encoded hidden states of the neighbouring trajectories in the local neighbourhood. \par
Hardwired attention weights are designed to incorporate the notion of distance between the pedestrian of interest and his or her neighbours into the trajectory prediction model. The closer a neighbouring pedestrian, the higher their associated weight, because that pedestrian has a greater influence on the trajectory that we are trying to predict.

The simplest representation scheme can be given by,
\begin{equation}
	w_{(n,j)}=\cfrac{1}{\mathrm{dist}(n,j)}  ,
\end{equation}
where $\mathrm{dist}(n,j)$ is the distance between the $n^{th}$ neighbour and the pedestrian of interest at the $j^{th}$ time instance, and $w_{(n,j)}$ is the generated hardwired attention weight. This idea can be extended to generate the context vector for the hardwired attention model. \par
Let there be $N$ neighbouring trajectories in the local neighbourhood and $h^{'}_{(n,j)}$ be the encoded hidden state of the $n^{th}$ neighbour at the $j^{th}$ time instance, then the context vector for the hardwired attention model is defined as,
\begin{equation}
C^{h}=\sum_{n=1}^{N}\sum_{j=1}^{T_{obs}} w_{(n,j)}h^{'}_{(n,j)} .
\end{equation}

 We then employ a simple concatenation layer to combine the information from individual attentions. Hence the combined context vector can be denoted as, 
\begin{equation}
C_{t}^{*}=\mathrm{tanh}(W_c[C^{s}_{t};C^{h}]) ,
\end{equation}
where $W_c$ is referred to as the set of weights for concatenation. We learn this weight value also through back-propagation.\par

The final prediction can now be computed as,
\begin{equation}
\mathbf{y}_t=\mathrm{LSTM}(s_{t-1},\mathbf{y}_{t-1},C_{t}^{*}) ,
\end{equation}
where the decoding function $f$ in Eq. \ref{eq:4} is replaced with a LSTM decoder as we are employing LSTMs for encoding and decoding purposes. 

\subsection{Model Learning}
 The given input trajectories in the training set are clustered based on source and sink positions and we run an outlier detection algorithm for each cluster considering the entire trajectory. 
 For clustering we used DBSCAN (\cite{dbscan}) as it enables us to cluster the data on the fly without specifying the number of clusters. Hyper parameters of the DBSCAN algorithm were chosen experimentally.
 \par 
 In the training phase trajectories are clustered based on the entire trajectory and after clustering we learnt a separate trajectory prediction model for each generated cluster. When modelling the local neighbourhood of the pedestrians of interest, we have encoded the trajectories of those closest 10 neighbours in each direction, namely front, left and right. If there exist more than 10 neighbours in any direction, we have taken the first (closest) 9 trajectories and the mean trajectory of the rest of the neighbours. If a trajectory has less than 10 neighbours, we create dummy trajectories such that we have 10 neighbours, and set the weight of these dummy neighbours to 0.\par

 When testing the model, we are concerned with predicting the pedestrian trajectory given the first $T_{obs}$ locations. To select the appropriate prediction model to use, the mean trajectory for each cluster for the period of ${1}$ to ${T_{obs}}$ is generated and in the testing phase, the given trajectories are assigned to the closest cluster centre while considering those mean trajectories as the cluster centroids. 
 
\subsection{Comparison to the Social-LSTM model of \citet{social_LSTM}}
In this section we draw comparisons between the current state-of-the-art technique and the proposed approach.
In \cite{social_LSTM}, for each neighbouring pedestrian, the hidden state at time $t-1$ is extracted out and fed as an input to the prediction model of the pedestrian of interest. Let there be $N$ neighbours in the local neighbourhood and $h^{'}_{(n,t-1)}$ be the hidden state of the $n^{th}$ neighbour at the time instance $t-1$. Then the process can be written as, 
\begin{equation}
H^{'}_t=\sum_{n=1}^{N}h^{'}_{(n,t-1)} ,
\end{equation}

and the hidden state of the pedestrian of interest at the $t^{th}$ time instance is given by, 
\begin{equation}
h_t=\mathrm{LSTM}(h_{t-1},\mathbf{x}_{t-1},H^{'}_t) ,
 \label{eq:social_lstm}
\end{equation}
where $h_{t-1}$ refers to the hidden state and $\mathbf{x}_{t-1}$ refers to the position of the pedestrian of interest at the $t-1$ time instance. The authors are passing $H^{'}_t$ and $\mathbf{x}_{t-1}$ through embedding functions before feeding it to the LSTM model, but in order to draw direct comparisons we are using the above notation. In \citet{social_LSTM}, the hidden state of the pedestrian of interest at the $t^{th}$ time instance depends only on his or her previous hidden state, the position in the previous time instance and the pooled hidden state of the immediately preceding time step for the neighbouring pedestrians (see Eq. \ref{eq:social_lstm}). Comparing to our model, we are considering the entire set of hidden states for the pedestrian of interest as well as the neighbouring pedestrians when predicting the $t^{th}$ output element (see Eq. 5-11). 
\par
As humans we tend to vary our intentions time to time. For an example consider the problem of navigating in a train station. A person may start walking towards the desired platform and then realise that he hasn't got a ticket and then make a sudden change and move towards the ticket counter. When applying the LSTM model proposed by \citet{social_LSTM} to such real world scenarios, by observing the immediate preceding hidden state one can generate reactive behaviour to avoid collisions but when doing long term path planning, even though the LSTM is capable of handling long term relationships, the prediction process may go almost ``blindly'' towards the end of the sequence (\cite{JiaGFT15}) as we are neglecting vital information about pedestrian's behaviour under varying contexts. In contrast, the proposed combined attention model considers the entire sequence of hidden states for both the pedestrian of interest and his or her neighbours and then we utilise time dependent weights which enables us to vary their influence in a timely manner. 
\par
Additionally, we observed that even in unstructured scenes such as train stations, airport terminals and shopping malls where multiple source and sink positions are present, still there exists dominant motion patterns describing the navigation preference of the pedestrians. For instance taking the same train station example, although the main problem that we are trying to solve here is to navigating while avoiding collisions, humans demonstrate different preferences in doing so. One pedestrian may be there to meet passengers and hence is wandering in a free area, while another pedestrian may aim to get in or out of the train station as quickly as possible. Therefore one single LSTM model is not sufficient to capture such ambiguities in navigational patterns. We observed that such distinct preferences in navigation generate unique trajectory patterns which can be easily segmented via the proposed clustering process. Therefore in contrast to \citet{social_LSTM}, we are learning a different trajectory prediction model for each trajectory cluster.

\section{Experiments}
\begin{figure*}[t!]
    \centering
     \begin{subfigure}[t]{0.45\textwidth}
        \centering
        \includegraphics[width=4.8cm,height=3.0cm,keepaspectratio]{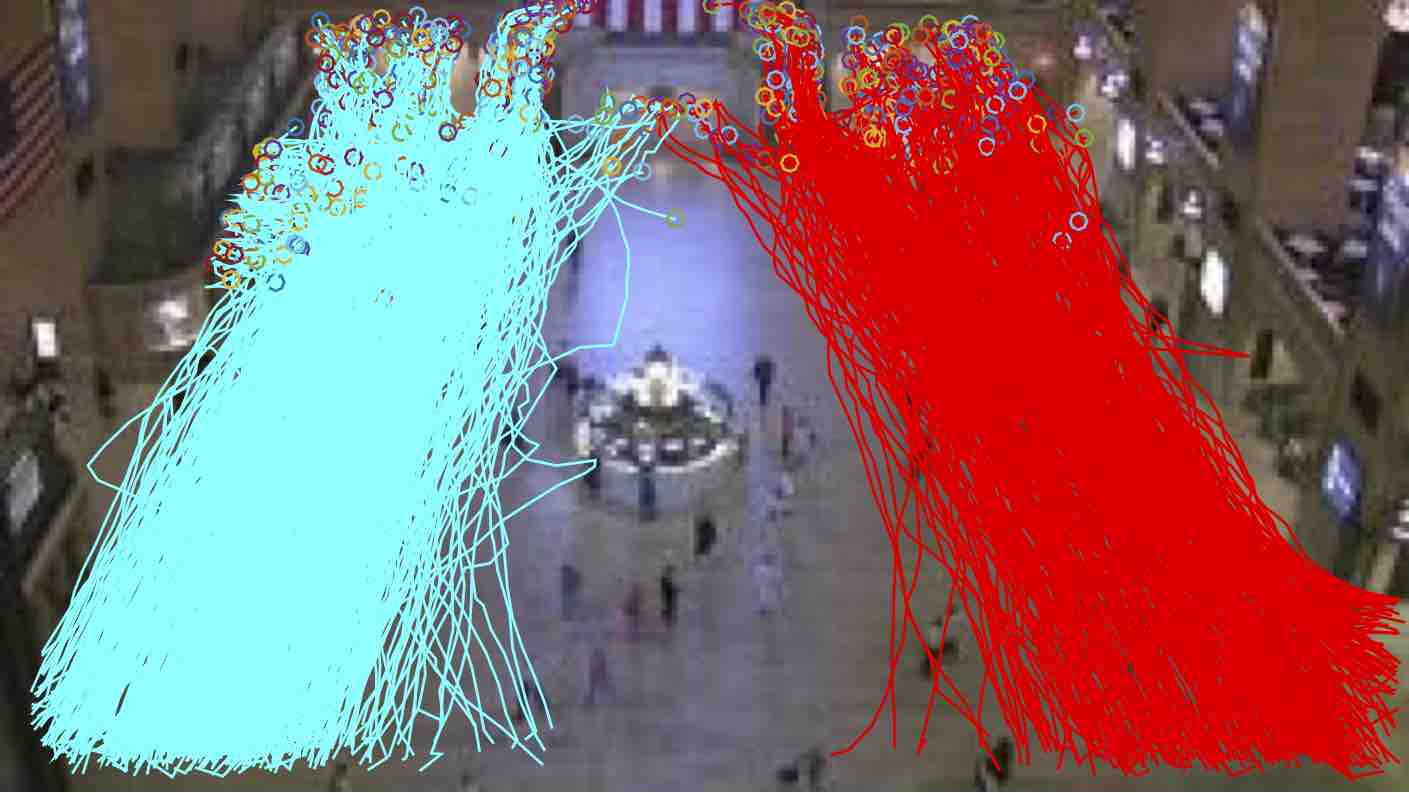} 
        \caption{First 2 clusters}
    \end{subfigure}%
    \begin{subfigure}[t]{0.45\textwidth}
        \centering
        \includegraphics[width=4.8cm,height=3.0cm,keepaspectratio]{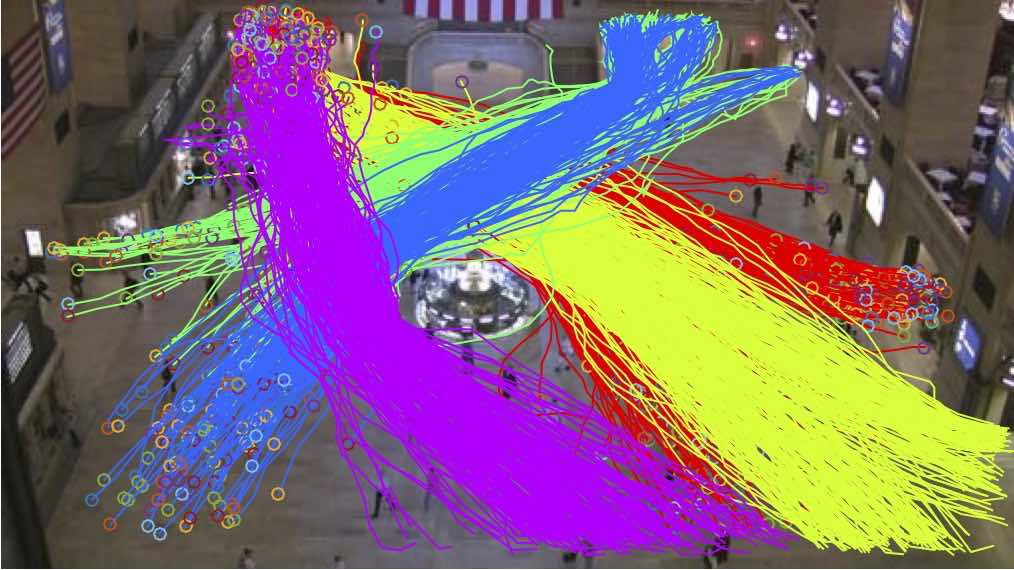}
        \caption{Next 5 clusters}
    \end{subfigure}
          \begin{subfigure}[t]{0.45\textwidth}
        \centering
        \includegraphics[width=4cm,height=3cm,keepaspectratio]{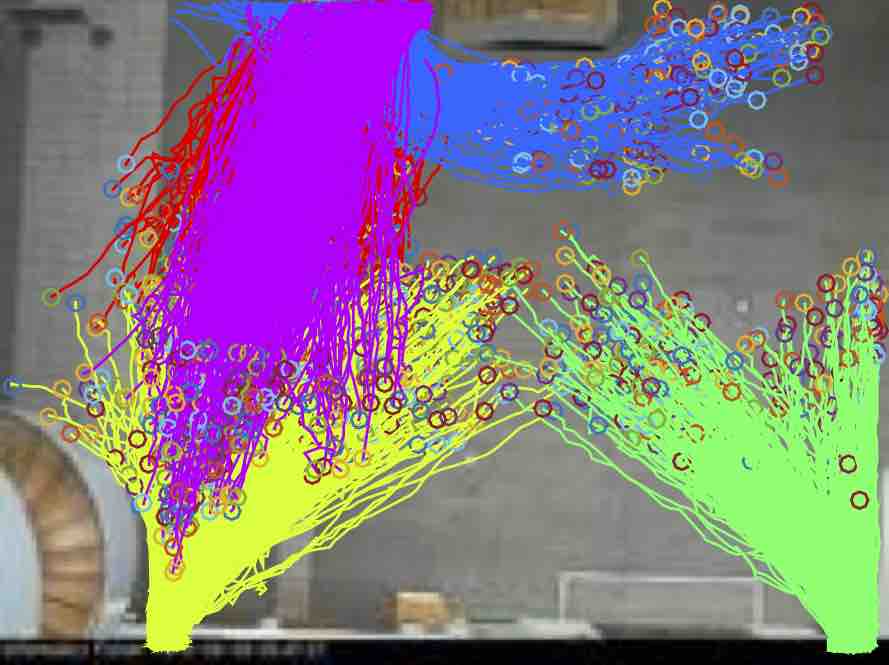} 
        \caption{First 5 clusters}
    \end{subfigure}
    \begin{subfigure}[t]{0.45\textwidth}
        \centering
        \includegraphics[width=4cm,height=3cm,keepaspectratio]{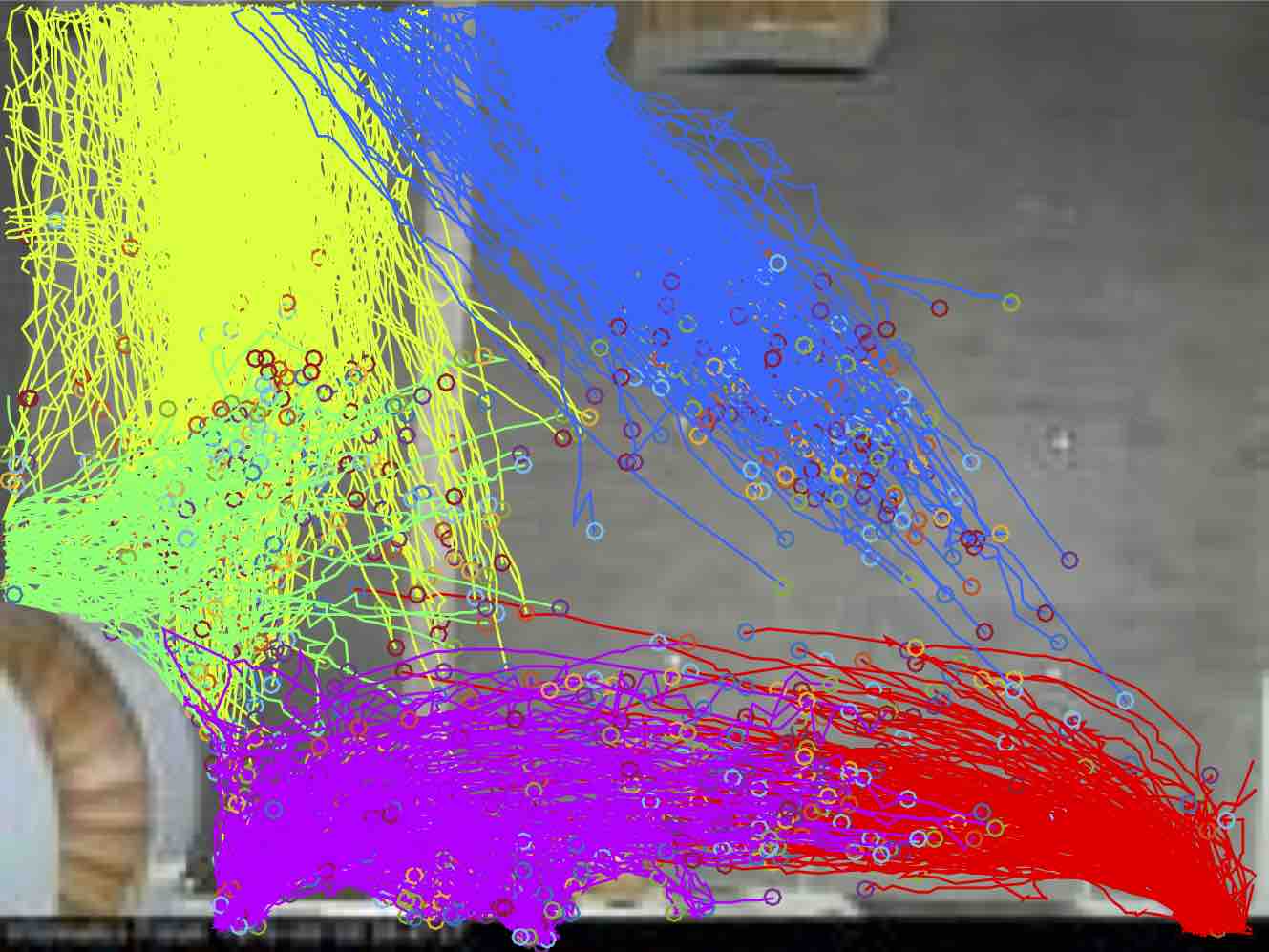}
        \caption{Next 5 clusters}
    \end{subfigure}
		\vspace{-1mm}		
    \caption{Clustering results for Grand Central (a, b) and Edinburgh Informatics Forum (c, d) Datasets}
        \label{fig:fig_clustering}
		
\end{figure*}

We present the experimental results on two publicly available human trajectory datasets: New York Grand Central (GC) (\citet{Yi_CVPR_2015}) and Edinburgh Informatics Forum (EIF) database (\citet{EIF}). The Grand Central dataset consist of around 12,600 trajectories where as the Edinburgh Informatics Forum database contains around 90,000 trajectories. We have conducted 2 experiments. For the first experiment on the Grand Central dataset, after filtering out short and fragmented trajectories \footnote{We consider short trajectories to be those with length less than the time period that we are considering  (40 frames) where as fragmented trajectories are trajectories which have discontinuities between 2 consecutive frames due to noise in the tracking process.}, we are left with 8,000 trajectories, and train our model on 5,000 trajectories and evaluate the prediction accuracy on 3,000 trajectories. In the next experiment we considered 3 days worth trajectories from Edinburgh Informatics Forum database, trained our model on 10,000 trajectories and tested on 5,000 trajectories. \par

Prior to learning trajectory models, we employ clustering to separate the different modes of human motion. This allows us to learn separate models for different behaviours, such as one model for a pedestrian who is buying tickets and another for those who are directly entering or leaving the train station. We believe that these different motion patterns generate unique pedestrian behavioural styles and that are well captured through separate models. This can be achieved via clustering trajectories based on entire trajectory, but this will produce very large number of clusters or large number of outliers due to the wide variation in the different modes of human motion. As a result, in each cluster, we would have very few examples to train our prediction models on. Therefore as a solution to the above stated problem we cluster the trajectories based only on the enter/exit zones. As illustrated in Fig. \ref{fig:fig_clustering} this approach works reasonably well at separating different modes of human motion. \par
Even with clustering based on entry and exit points, the way that a person moves through the environment and how they are influenced by neighbours will vary considerably. For instance consider the clusters represented in green and blue in Fig. \ref{fig:fig_clustering} (b). The way that an intersecting trajectory travelling straight up in the scene, from bottom left towards up left, will affect green and blue clusters differently because of their different exit zones. It's general human nature to try to avoid collisions while keeping the expected heading direction. Therefore entry/exit zones based clustering is sufficient to capture this. 

\subsection{Quantitative results}
Similar to \citet{social_LSTM} we report prediction accuracy with the following 3 error metrics. Let $n$ be the number of trajectories in the testing set, $\mathbf{x}_{i,t}^{pred}$ be the predicted position for the trajectory $i$ at $t^{th}$ time instance, and $\mathbf{x}_{i,t}^{obs}$ be the respective observed positions then,

\begin{enumerate}
\item \textit{Average displacement error (ADE): } 
\begin{equation}
ADE=\cfrac{\sum\limits_{i=1}^{n}\sum\limits_{t=T_{obs}+1}^{T_{pred}}(\mathbf{x}_{i,t}^{pred}-\mathbf{x}_{i,t}^{obs})^2}{n(T_{pred}-(T_{obs}+1))} .
\end{equation}
\item \textit{Final displacement error (FDE) :} 
\begin{equation}
FDE=\cfrac{\sum\limits_{i=1}^{n}\sqrt{(\mathbf{x}_{i,T_{pred}}^{pred}-\mathbf{x}_{i,T_{pred}}^{obs})^2}}{n} .
\end{equation}
\item \textit{Average non-linear displacement error (n-ADE): } The average displacement error for the non-linear regions of the trajectory,
\begin{equation}
n-ADE=\cfrac{\sum\limits_{i=1}^{n}\sum\limits_{t=T_{obs}+1}^{T_{pred}}I(\mathbf{x}_{i,t}^{pred})(\mathbf{x}_{i,t}^{pred}-\mathbf{x}_{i,t}^{obs})^2}{\sum\limits_{i=1}^{n}\sum\limits_{t=T_{obs}+1}^{T_{pred}}I(\mathbf{x}_{i,t}^{pred})} ,
\end{equation}
where,
\begin{equation}
I(\mathbf{x}_{i,t}^{pred}) = \begin{cases} 1 &\mbox{if }  \cfrac{d^2y_{i,t}}{dx_{i,t}^2} \neq 0 .\\ 
0 & o. w \end{cases}
\end{equation}

\end{enumerate} 

In all experiments we have observed the trajectory (and it's neighbours) for 20 frames and predicted the trajectory for the next 20 frames. Compared to \cite{social_LSTM}, which has considered sequences of 20 frames total length, we are considering more lengthy sequences (with a total of 40 frames) as in \citet{Baccouche:2011} the authors have shown that LSTM models tend to generate more accurate results with lengthy sequences. \par
In the experimental results, tabulated in Tab \ref{tab:tab_1}, we compare our prediction model with the state-of-the-art. As the baseline models we implemented Social Force (SF) model from \citet{conf/cvpr/YamaguchiBOB11} and Social LSTM (S-LSTM) model given in \citet{social_LSTM}. For S-LSTM  model a local neighbourhood of size 32px was considered and the hyper-parameters were set according to \citet{social_LSTM}. In order to make direct comparisons with \citet{social_LSTM}, the hidden state dimensions of encoders and decoders of all OUR models were set to be 300 hidden units.
  
For the SF model, preferred speed, destination, and social grouping factors are used to model the agent behaviour. When predicting the destination, a linear support vector machine was trained with the ground truth destination areas detected in the Sec. 3.2. \par
In order to evaluate the strengths of the proposed model, we compare this combined attention model $\mathrm{(OUR_{cmb})}$ and two variations on our proposed approach: 1) $\mathrm{OUR_{sft}}$, which ignore the neighbouring trajectories and considers only the soft attention component derived from the trajectory of the person of interest when making predictions; and 2) $\mathrm{OUR_{sc}}$ which omits the clustering stage such that only a single model (using combined attention weights) is learnt. 
\begin{table}[!h]
  \centering
  \begin{adjustbox}{width=.8\linewidth,center}
  \begin{tabular}{|c|c|c|c|c|c|c|}
    \hline
   Metirc & Dataset & SF & S-LSTM & $\mathrm{OUR_{sc}}$ & $\mathrm{OUR_{sft}}$ & $\mathrm{OUR_{cmb}}$ \\
    \hline
    \multirow{2}{*}{ADE} & GC & 3.364 & 1.990 &1.878 &2.041 & \textbf{1.096} \\
     \hhline{~------} &EIF  & 3.124 & 1.524 &1.392 & 1.685 & \textbf{0.986} \\
   
    \hline\hline
    \multirow{2}{*}{FDE} & GC & 5.808 & 4.519 & 4.317 &5.277 & \textbf{3.011} \\
    \hhline{~------}            &EIF  & 3.909 & 2.510 & 2.345 &3.089 & \textbf{1.311} \\
    \hline\hline
     \multirow{2}{*}{n-ADE} & GC & 3.983 & 1.781 & 1.701 & 2.304 & \textbf{0.985} \\
    \hhline{~------}           &EIF  & 3.394 & 2.398 & 2.098 &2.415 & \textbf{0.901} \\

    \hline
  \end{tabular}
  \end{adjustbox}
  \caption{Quantitative results. In all the methods forecast trajectories are of length 20 frames. The first 2 rows represents the Average displacement error, rows 3 to 4 are for Final displacement error and the final 2 rows are for Average non-linear displacement error.}
  \label{tab:tab_1}
\end{table}

\begin{figure*}[!t]
    \centering
    \begin{subfigure}[t]{0.3\textwidth}
        \centering
        \includegraphics[width=.95\textwidth]{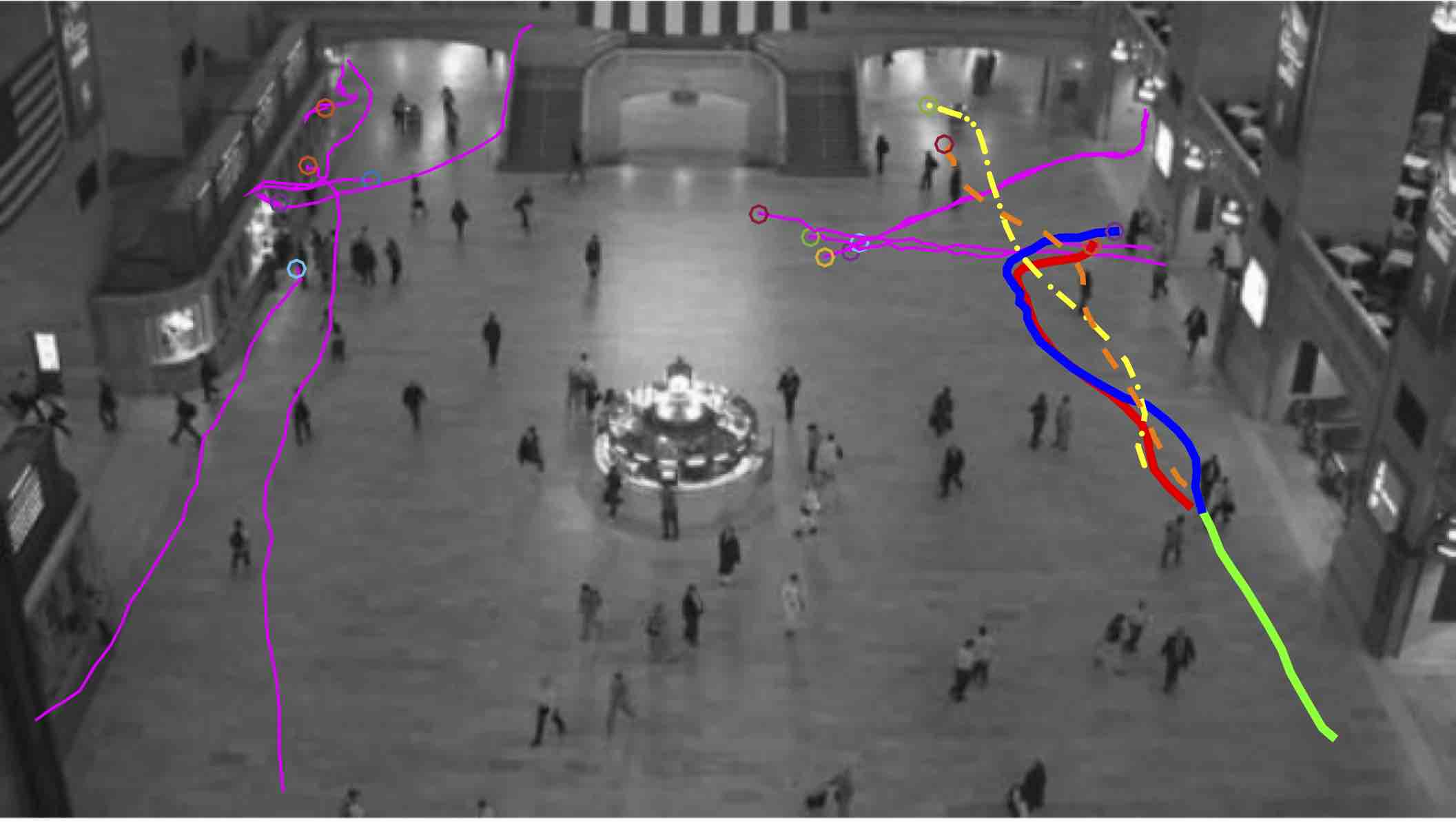} 
        \caption{}
    \end{subfigure}%
    \begin{subfigure}[t]{0.3\textwidth}
        \centering
        \includegraphics[width=.95\textwidth]{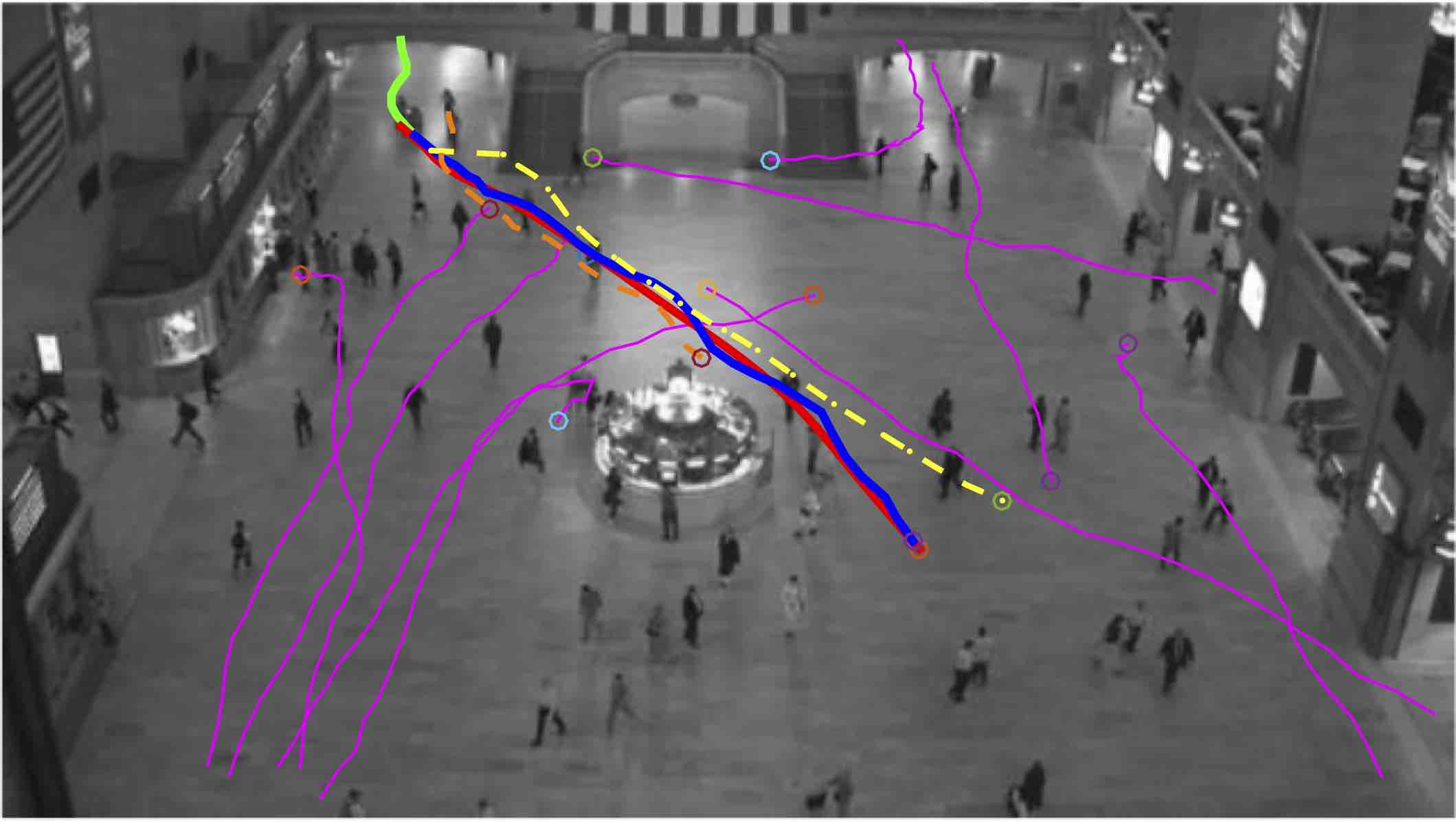} 
        \caption{}
    \end{subfigure}
     \begin{subfigure}[t]{0.3\textwidth}
        \centering
        \includegraphics[width=.95\textwidth]{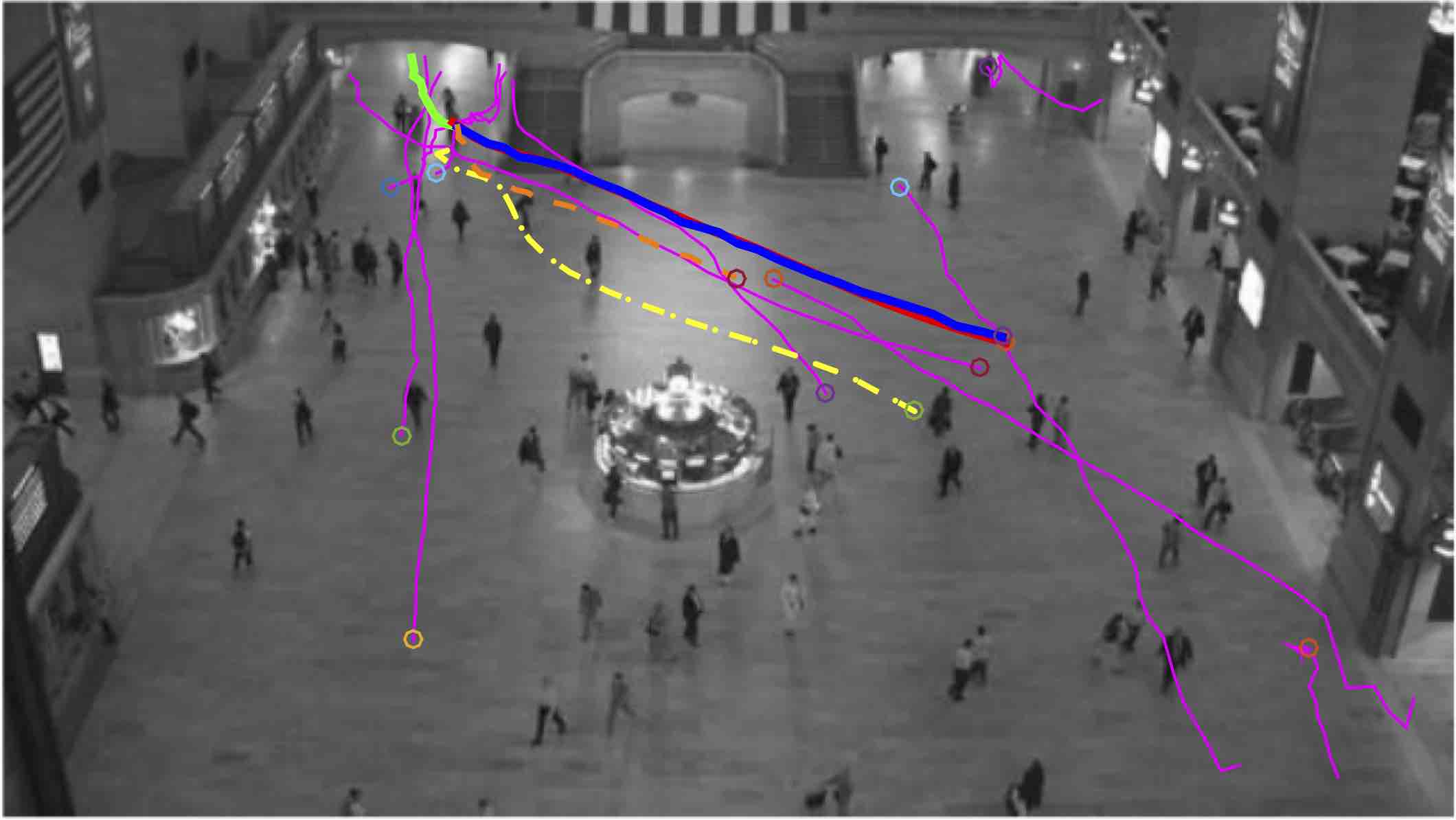}
        \caption{}
    \end{subfigure}
		\vspace{-1mm}
    \begin{subfigure}[t]{0.3\textwidth}
        \centering
        \includegraphics[width=.95\textwidth]{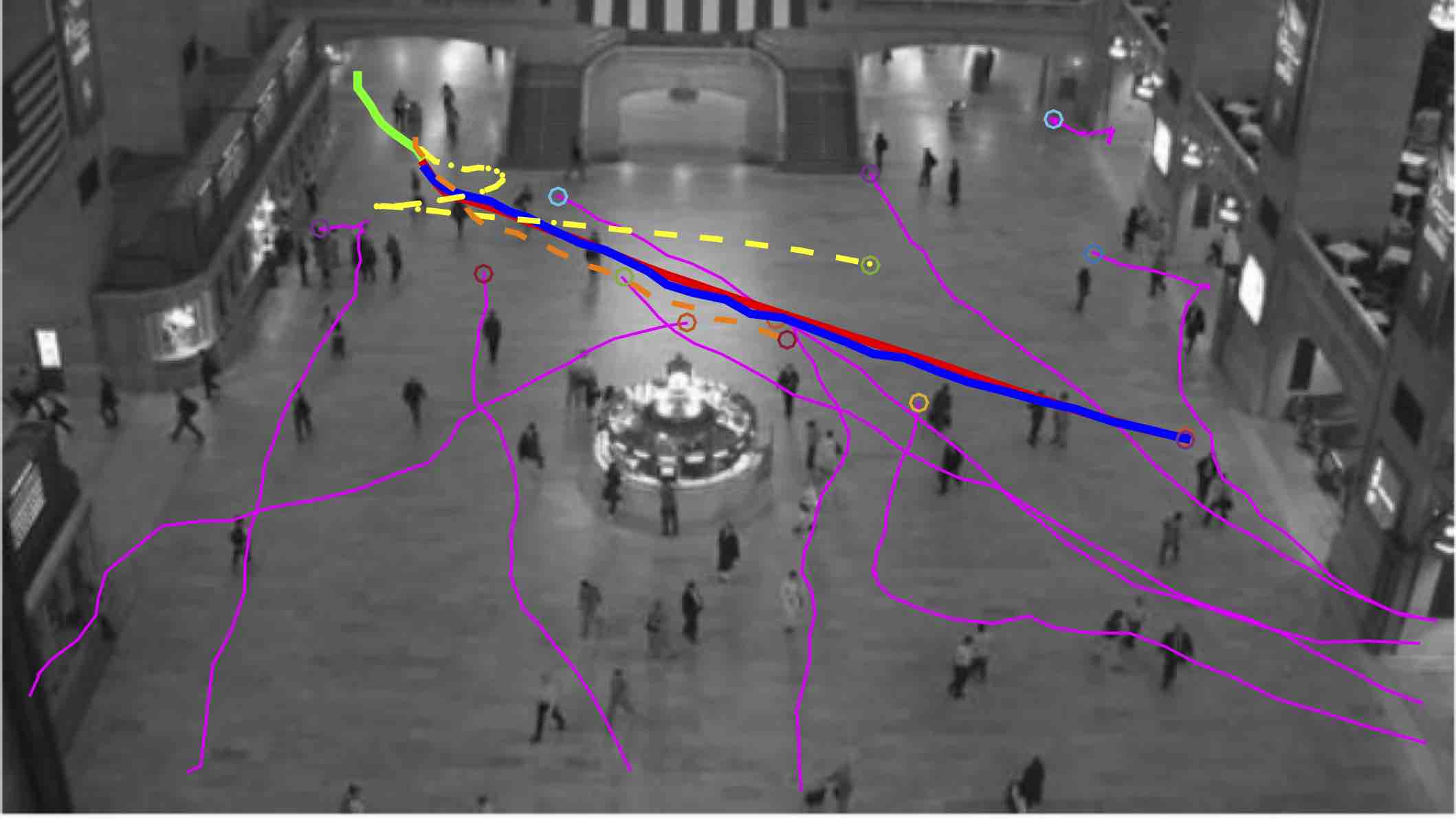}
        \caption{}
    \end{subfigure}%
    \begin{subfigure}[t]{0.3\textwidth}
        \centering
        \includegraphics[width=.95\textwidth]{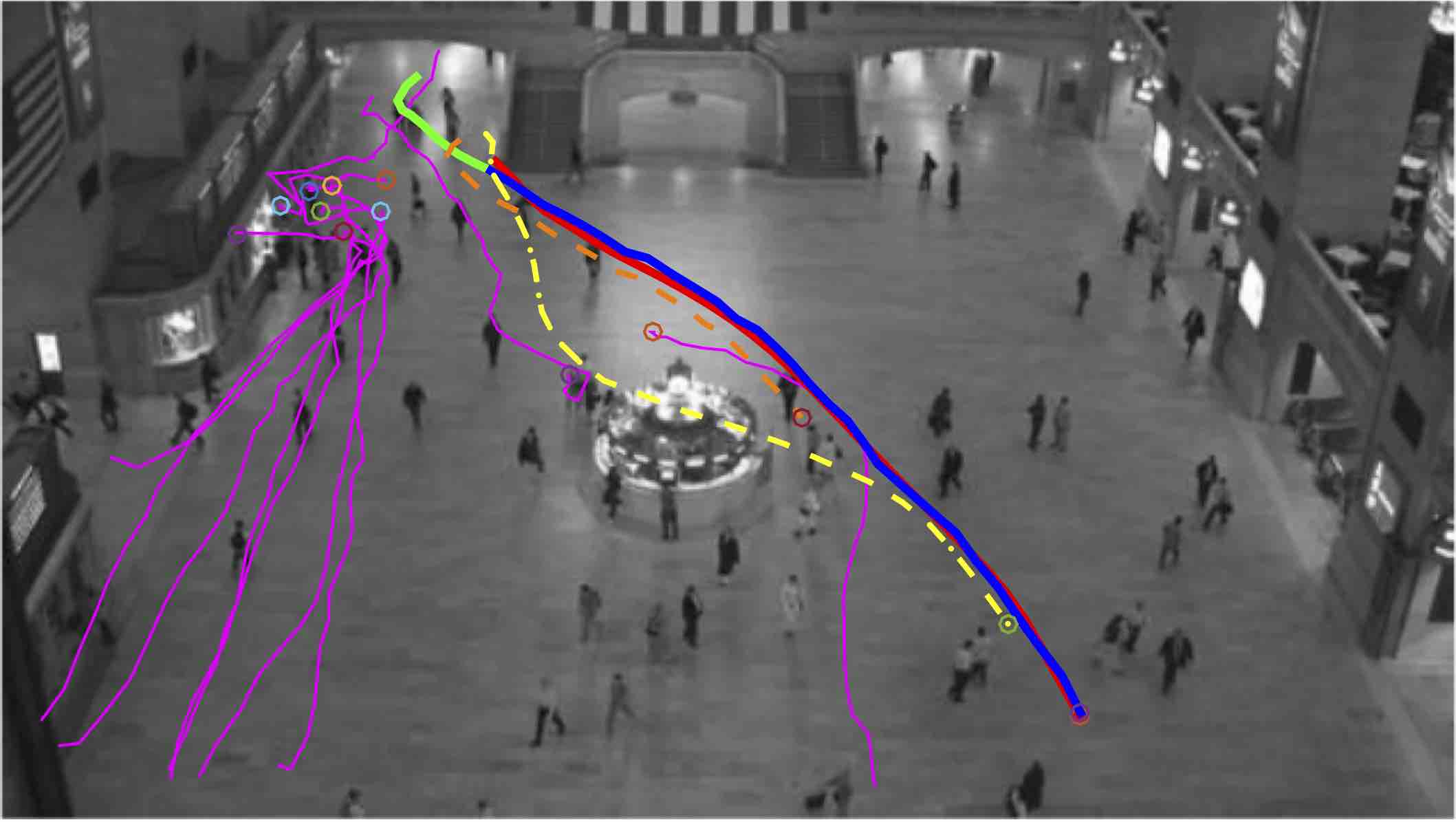}
        \caption{}
    \end{subfigure}
     \begin{subfigure}[t]{0.3\textwidth}
        \centering
        \includegraphics[width=.95\textwidth]{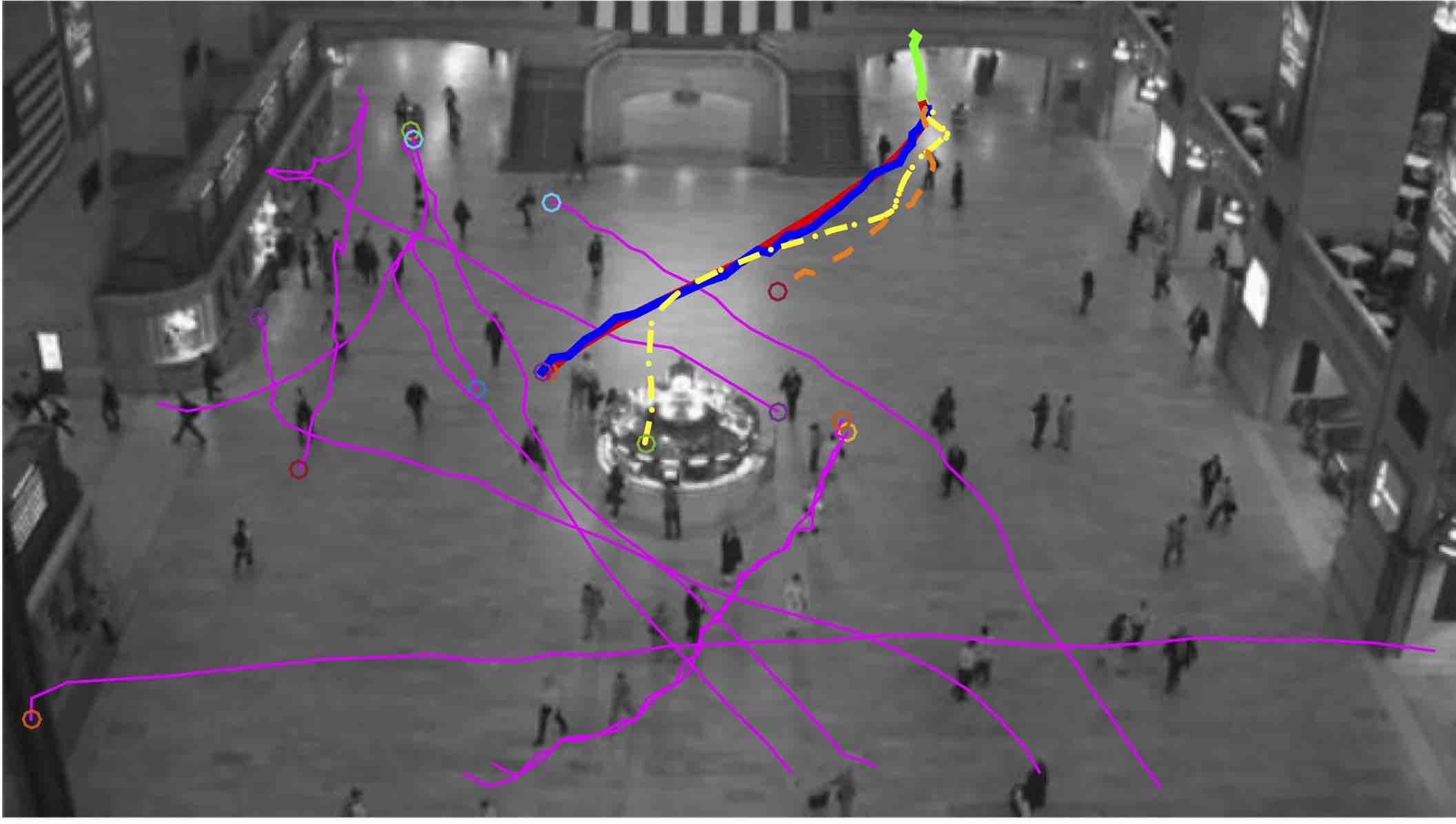}
        \caption{}
    \end{subfigure}
		\vspace{-1mm}
    \begin{subfigure}[t]{0.3\textwidth} 
        \centering
        \includegraphics[width=.95\textwidth]{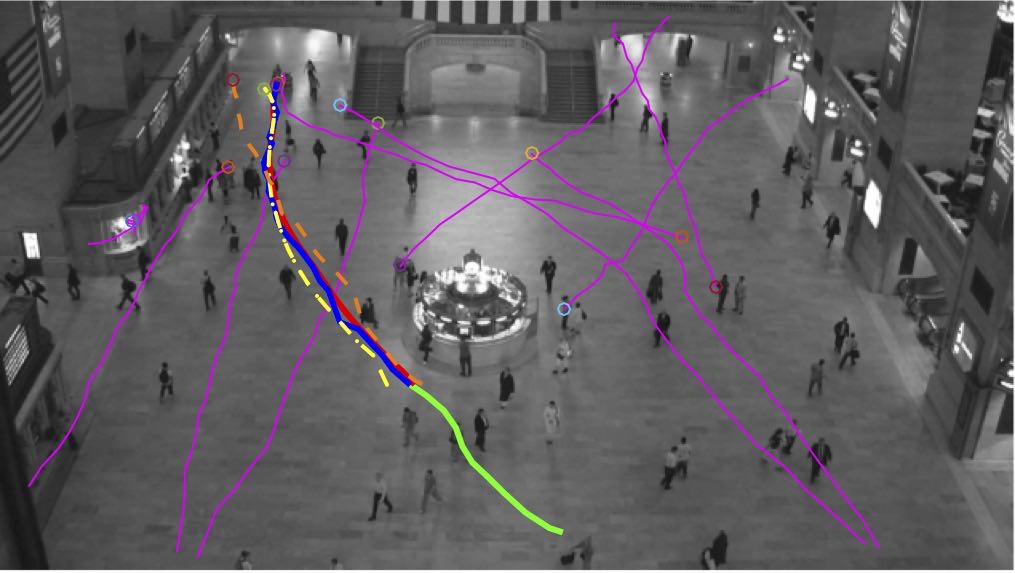}
        \caption{}
    \end{subfigure}%
    \begin{subfigure}[t]{0.3\textwidth}
        \centering
        \includegraphics[width=.95\textwidth]{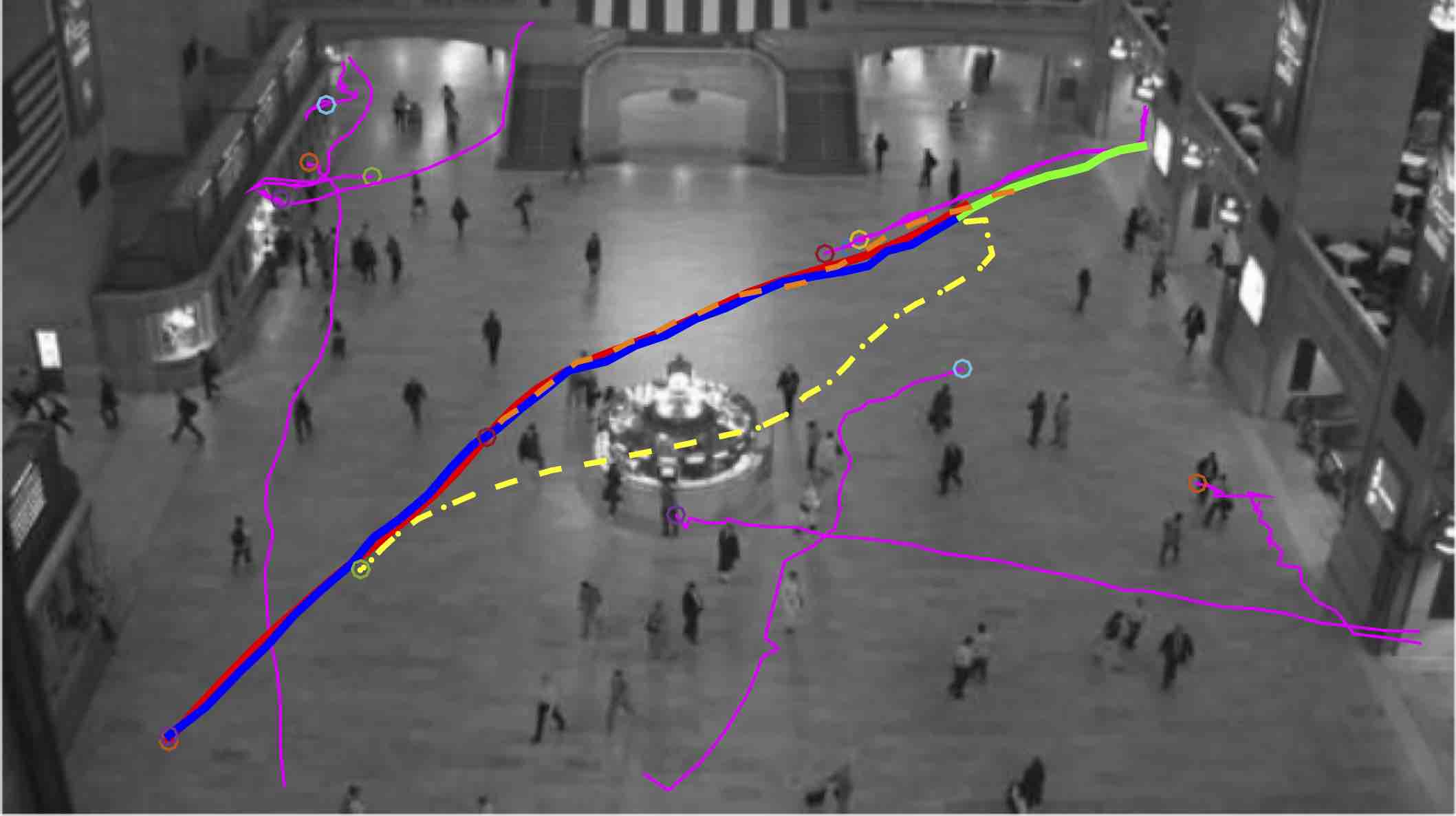}
        \caption{}
    \end{subfigure}
     \begin{subfigure}[t]{0.3\textwidth}
        \centering
        \includegraphics[width=.95\textwidth]{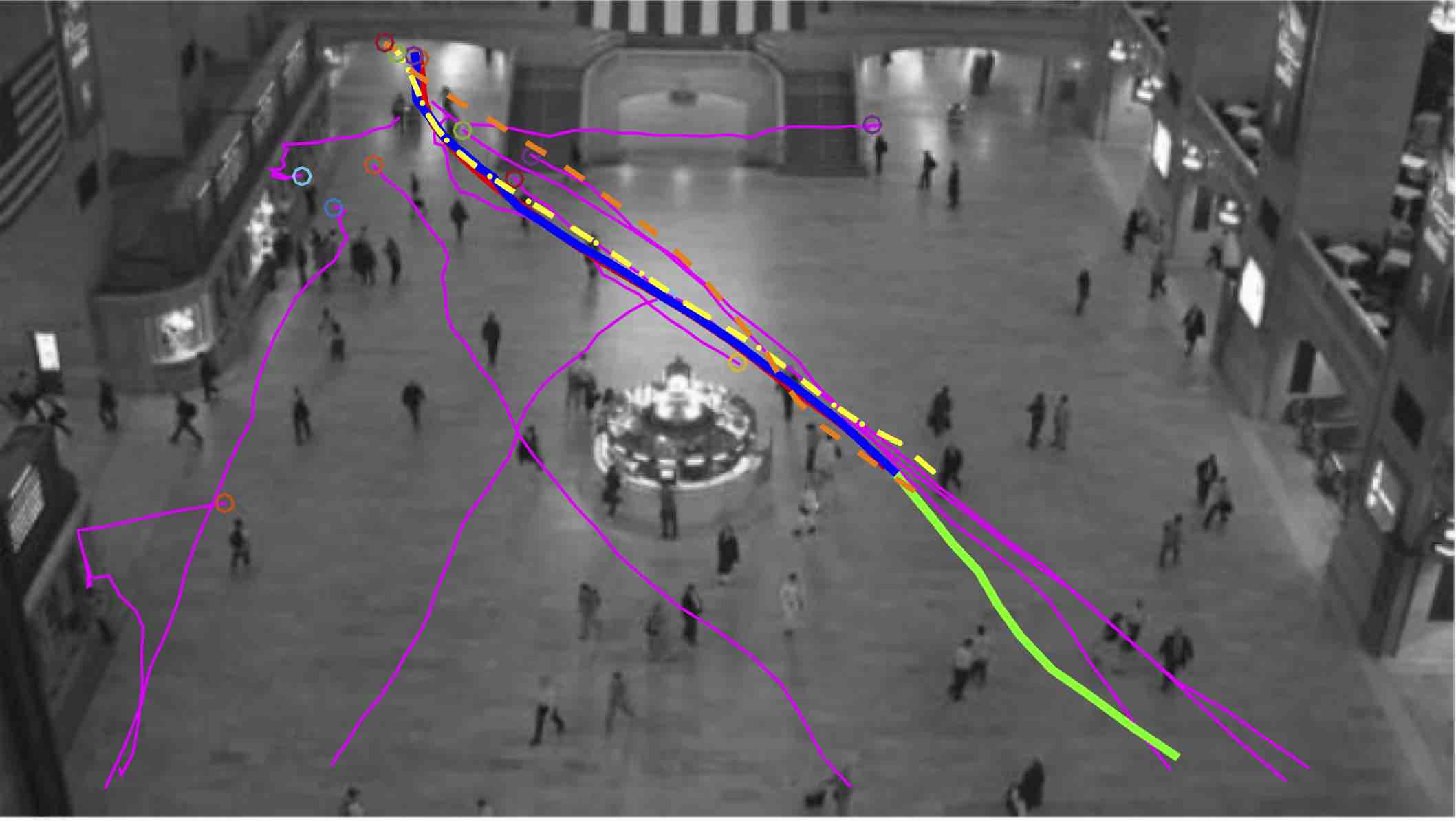}
        \caption{}
    \end{subfigure}
    \begin{subfigure}[t]{0.3\textwidth}
        \centering
        \includegraphics[width=.95\textwidth]{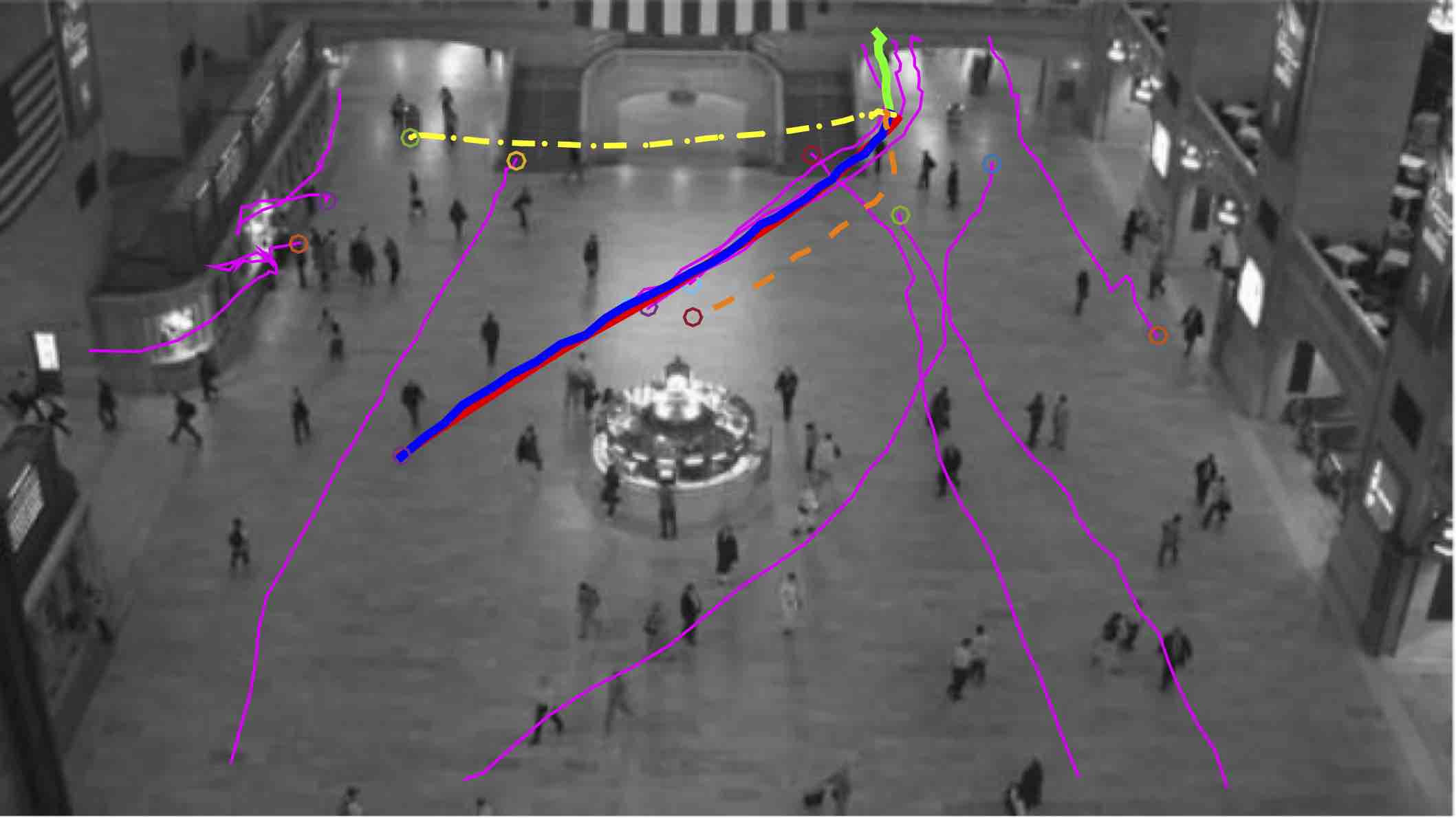}
        \caption{}
    \end{subfigure}%
    \begin{subfigure}[t]{0.3\textwidth}
        \centering
        \includegraphics[width=.95\textwidth]{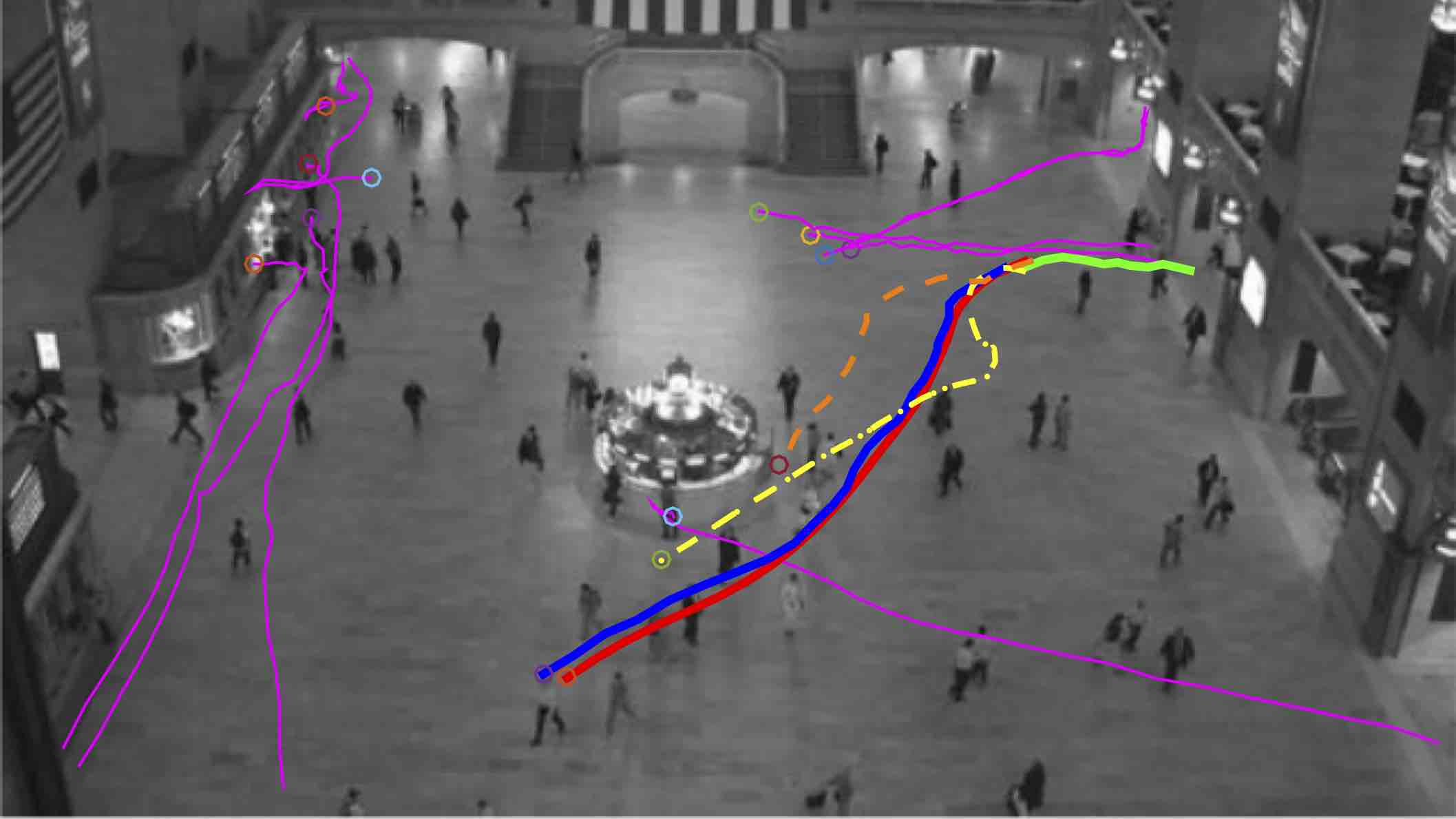}
        \caption{}
    \end{subfigure}
     \begin{subfigure}[t]{0.3\textwidth}
        \centering
        \includegraphics[width=.95\textwidth]{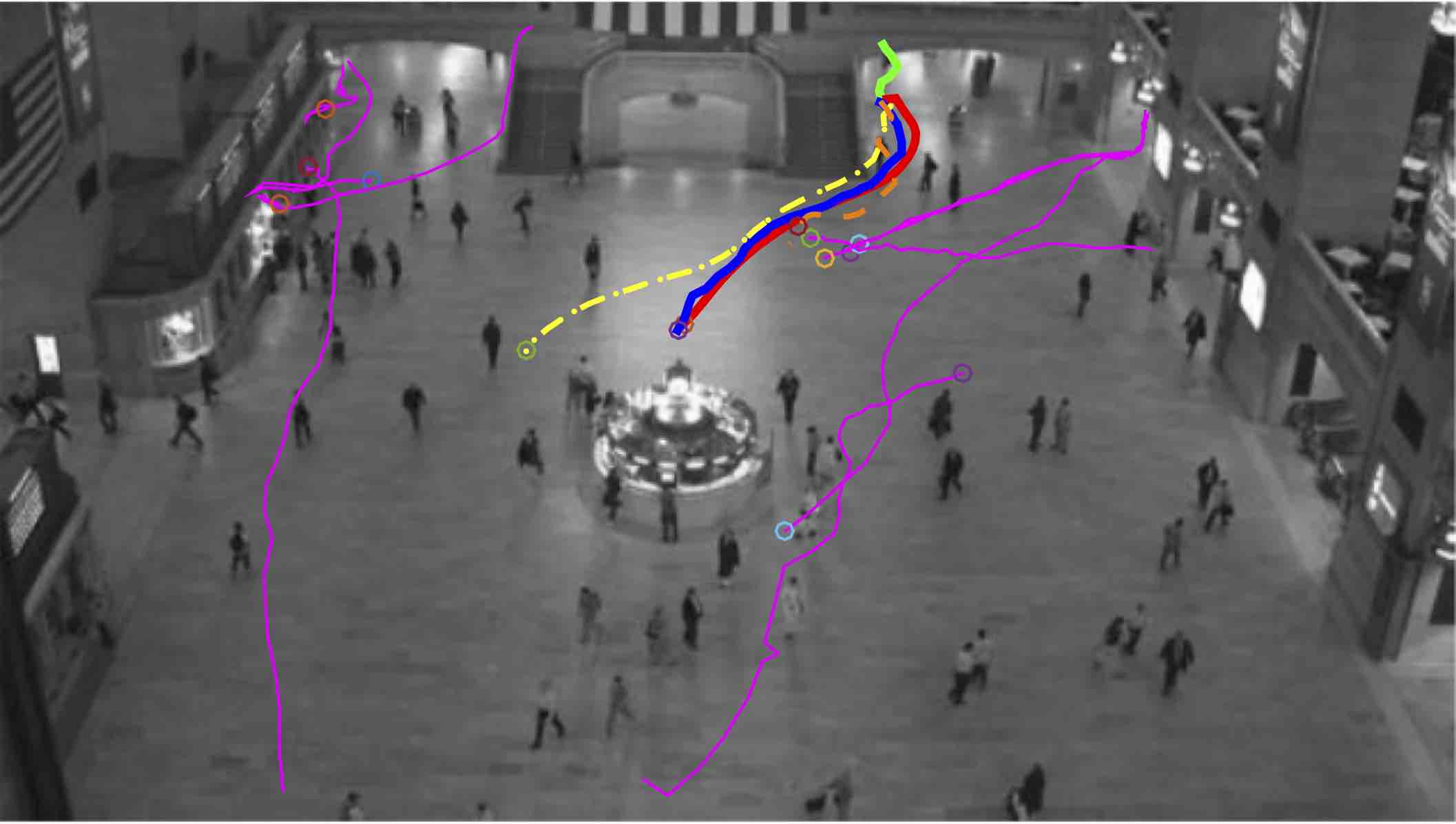}
        \caption{}
    \end{subfigure} 
     \begin{subfigure}[t]{0.3\textwidth}
        \centering
        \includegraphics[width=.95\textwidth]{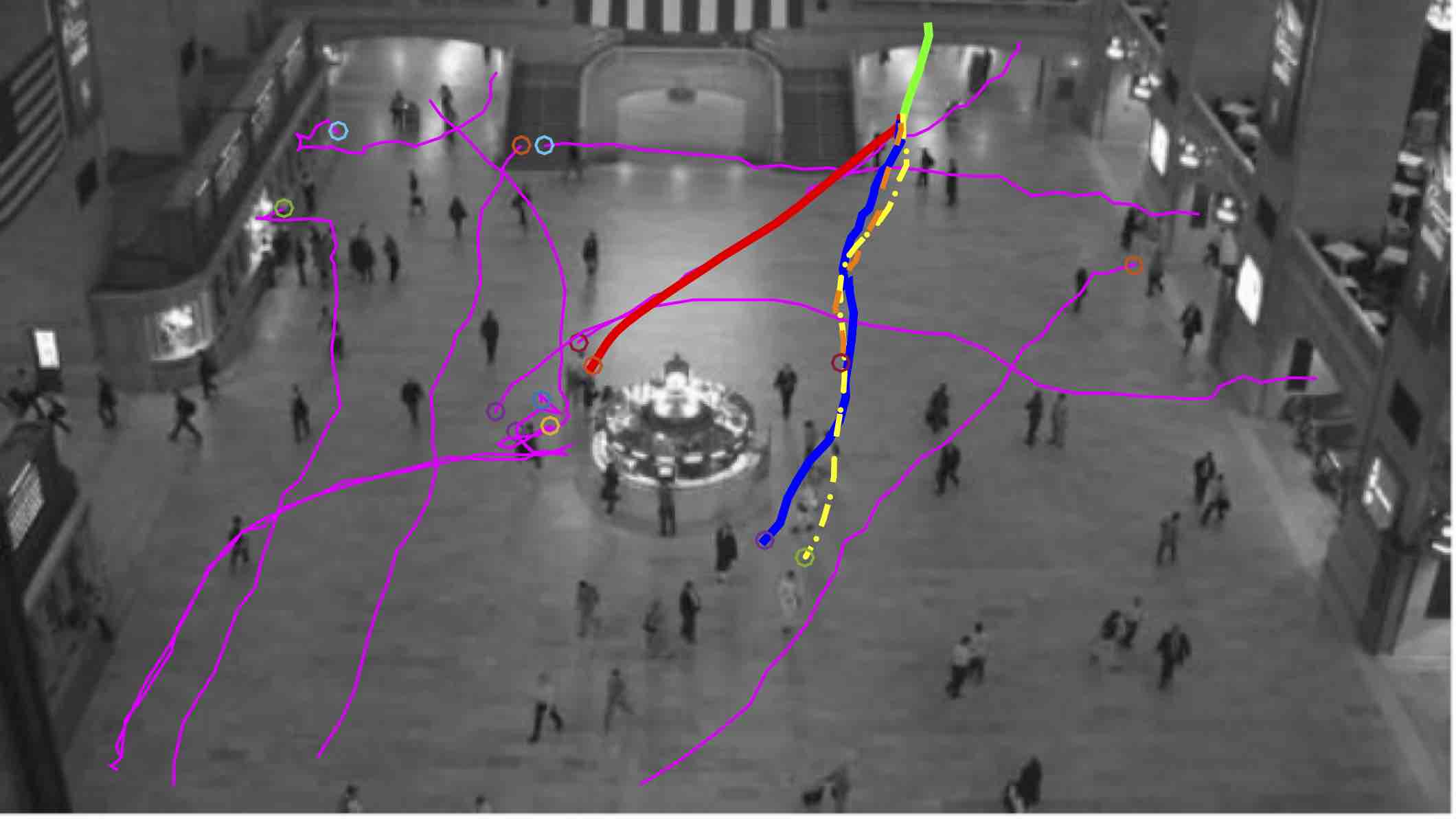}
        \caption{}
    \end{subfigure}%
    \begin{subfigure}[t]{0.3\textwidth}
        \centering
        \includegraphics[width=.95\textwidth]{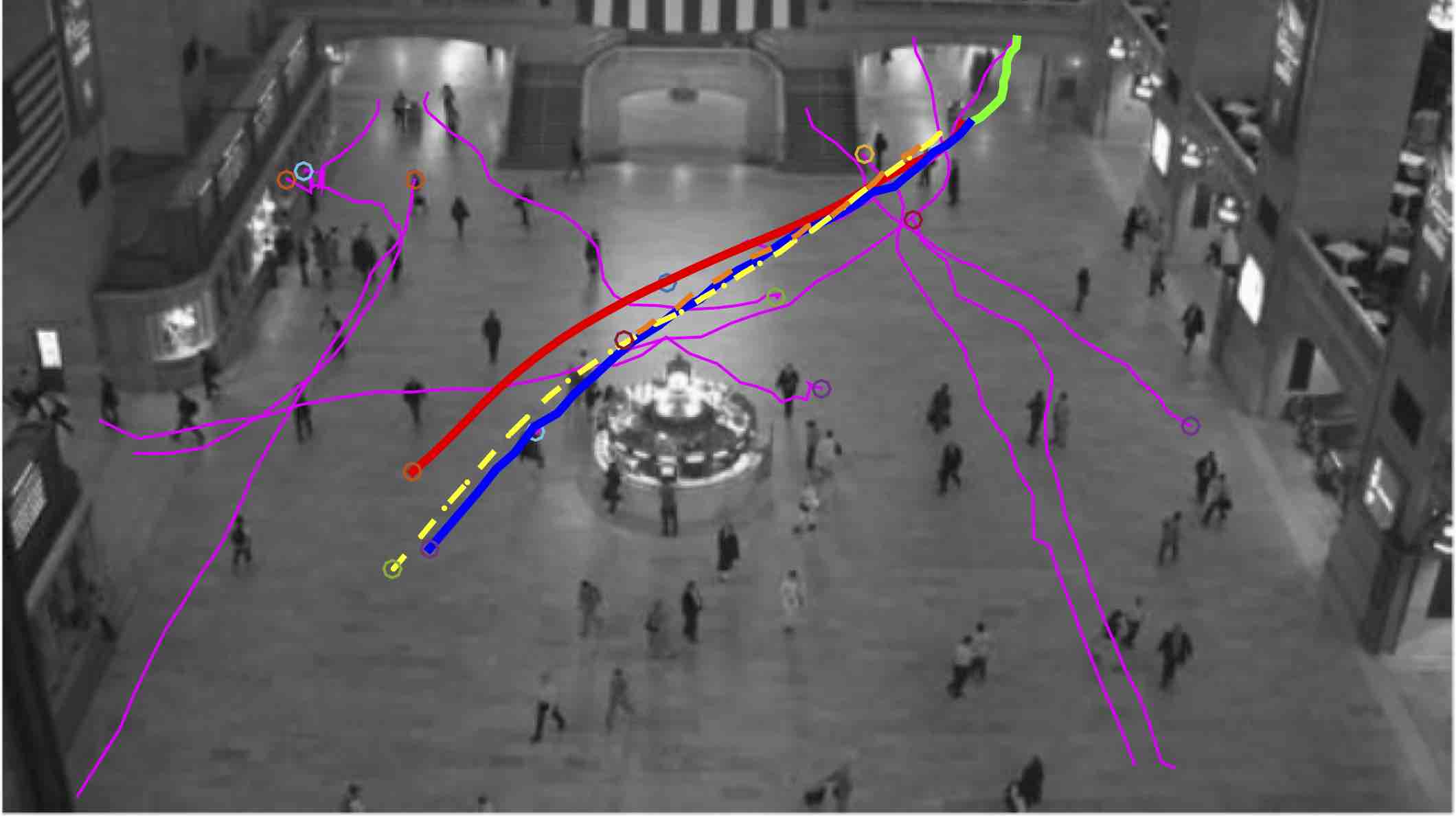}
        \caption{}
    \end{subfigure}
     \begin{subfigure}[t]{0.3\textwidth}
        \centering
        \includegraphics[width=.95\textwidth]{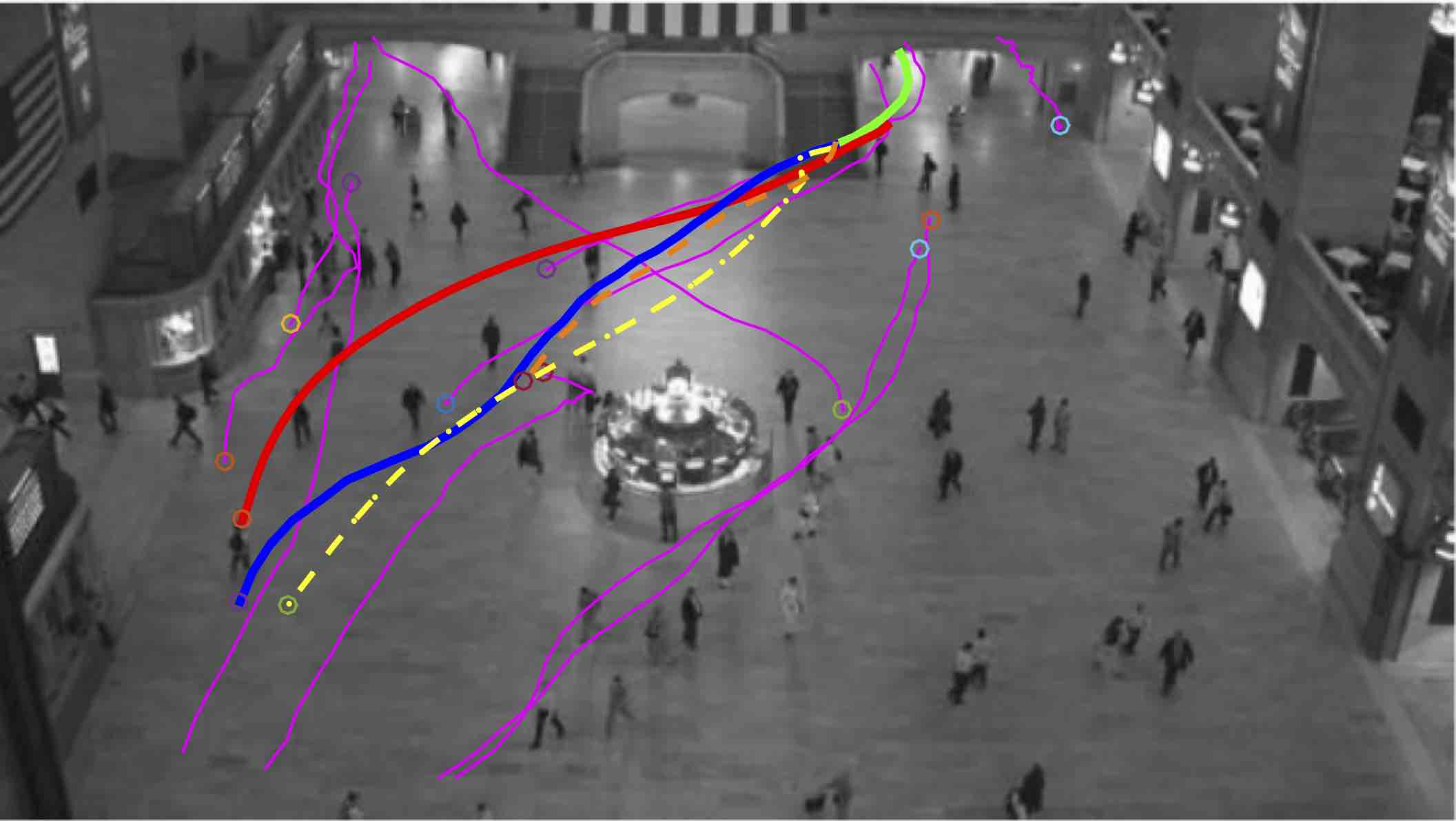}
        \caption{}
    \end{subfigure}
		\vspace{-2mm}
    \caption{Qualitative results: Given (in green), Ground Truth (in Blue), Neighbouring (in purple) and Predicted trajectories from $\mathbf{OUR_{cmb}}$ model (in red), from \textbf{S-LSTM} model (in yellow), from \textbf{SF} model (in orange). (a) to (l) accurate predictions and (m) to (o) some erroneous predictions}
    \label{fig:qualitative_resutls}
\end{figure*}

The proposed model outperforms the SF model and S-LSTM model in both datasets. The error reduction is more evident in the GC dataset where there are multiple source and sink positions, different crowd motion patterns are present and motion paths are heavily crowded. Comparing the results of $\mathbf{OUR_{sc}}$ (proposed approach without clustering) against the S-LSTM model we can see that regardless of the clustering process the proposed combined attention architecture is capable of improving the trajectory prediction. For all the measured error metrics $\mathbf{OUR_{sc}}$ has outperformed S-LSTM and $\mathrm{(OUR_{sft})}$ verifying that it is important to preserve historical data for both the pedestrian of interest as well as the neighbours. Secondly those results show that hard wired weights act as a good approximation of neighbours influence. \par When comparing results of $\mathbf{OUR_{cmb}}$ against $\mathbf{OUR_{sc}}$ it is evident that the clustering process has partitioned the trajectories based on these different semantics and via utilising separate models for each cluster, we were able to generate more accurate predictions. The combined attention model was capable of learning how the neighbours influence the current trajectory and how this impact varies under different neighbourhood locations. \par Furthermore, we would like to point out that, because of the separate model learning process we were able to predict the final destination positions with more precision compared to baseline models where they do not consider the environmental configurations of the unstructured scene.  While the proposed approach does not explicitly model the environment, a certain amount of environmental information is inherently encoded in the entry and exit locations, which the proposed approach is able to leverage. 
\subsection{Qualitative results}
In Fig. \ref{fig:qualitative_resutls} we show prediction results of the S-LSTM model, SF model and our combined attention model $\mathrm{(OUR_{cmb})}$ on the GC dataset. It should be noted that our model generates better predictions in heavily crowded areas. As we are learning a separate model for each cluster, the prediction models are able to learn different patterns of influences from neighbouring pedestrians. For instance in the 1st and 3rd column we demonstrate how the model adapts in order to avoid collisions.  In the last row of Fig. \ref{fig:qualitative_resutls} we show some failure cases. The reason for such deviations from the ground truth were mostly due to sudden changes in destination. Even though these trajectories do not match the ground truth, the proposed method still generates plausible trajectories. For instance, in  Fig. \ref{fig:qualitative_resutls} (m) and Fig. \ref{fig:qualitative_resutls} (o) the model moves side ways to avoid collusion with the neighbours in the left and right directions.  

\begin{figure*}[!t]
    \centering
    \begin{subfigure}[t]{0.23\textwidth}
        \centering
        \includegraphics[width=.95\textwidth]{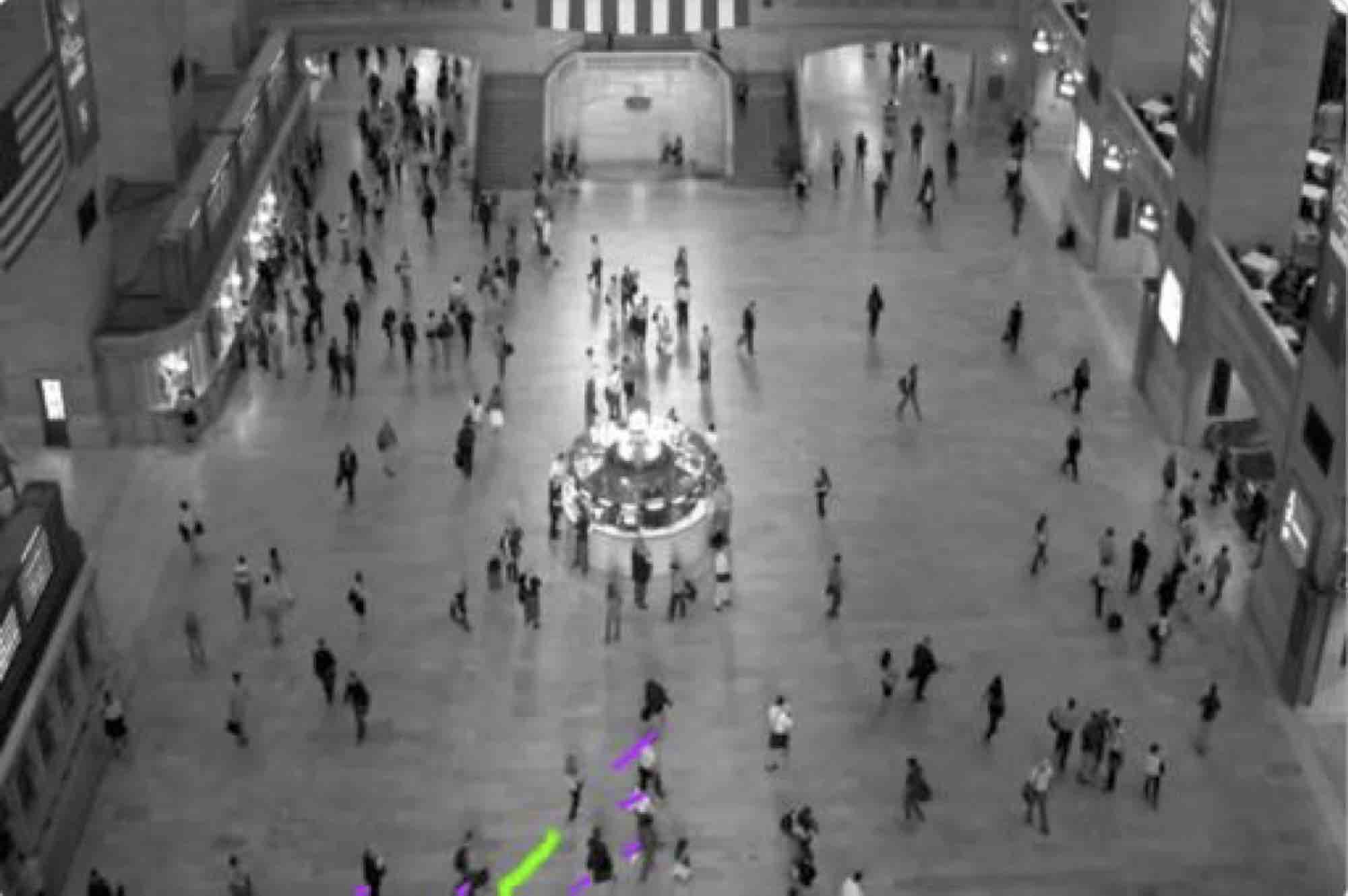} 
        \caption{}
    \end{subfigure}%
    \begin{subfigure}[t]{0.23\textwidth}
        \centering
        \includegraphics[width=.95\textwidth]{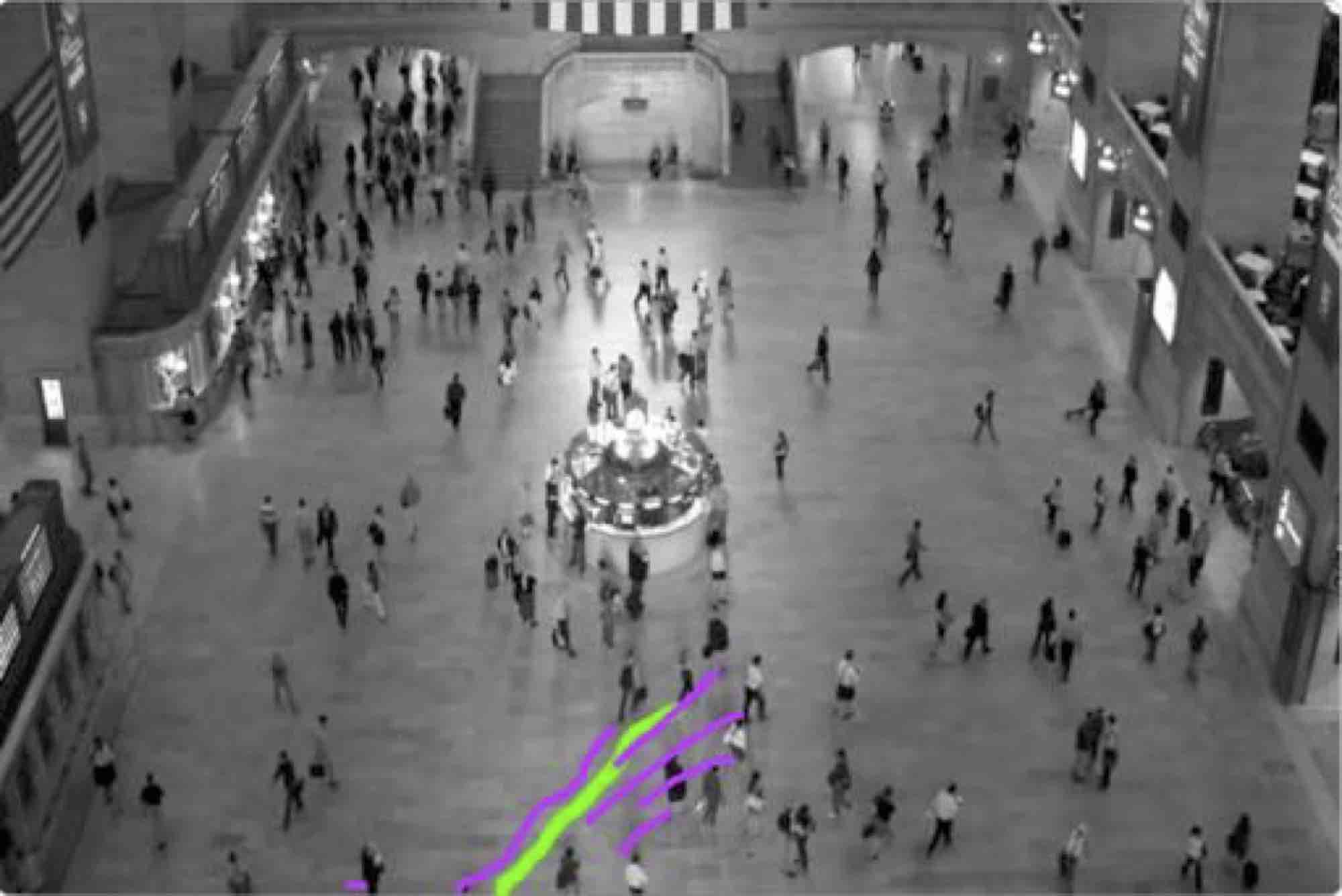} 
        \caption{}
    \end{subfigure}
     \begin{subfigure}[t]{0.23\textwidth}
        \centering
        \includegraphics[width=.95\textwidth]{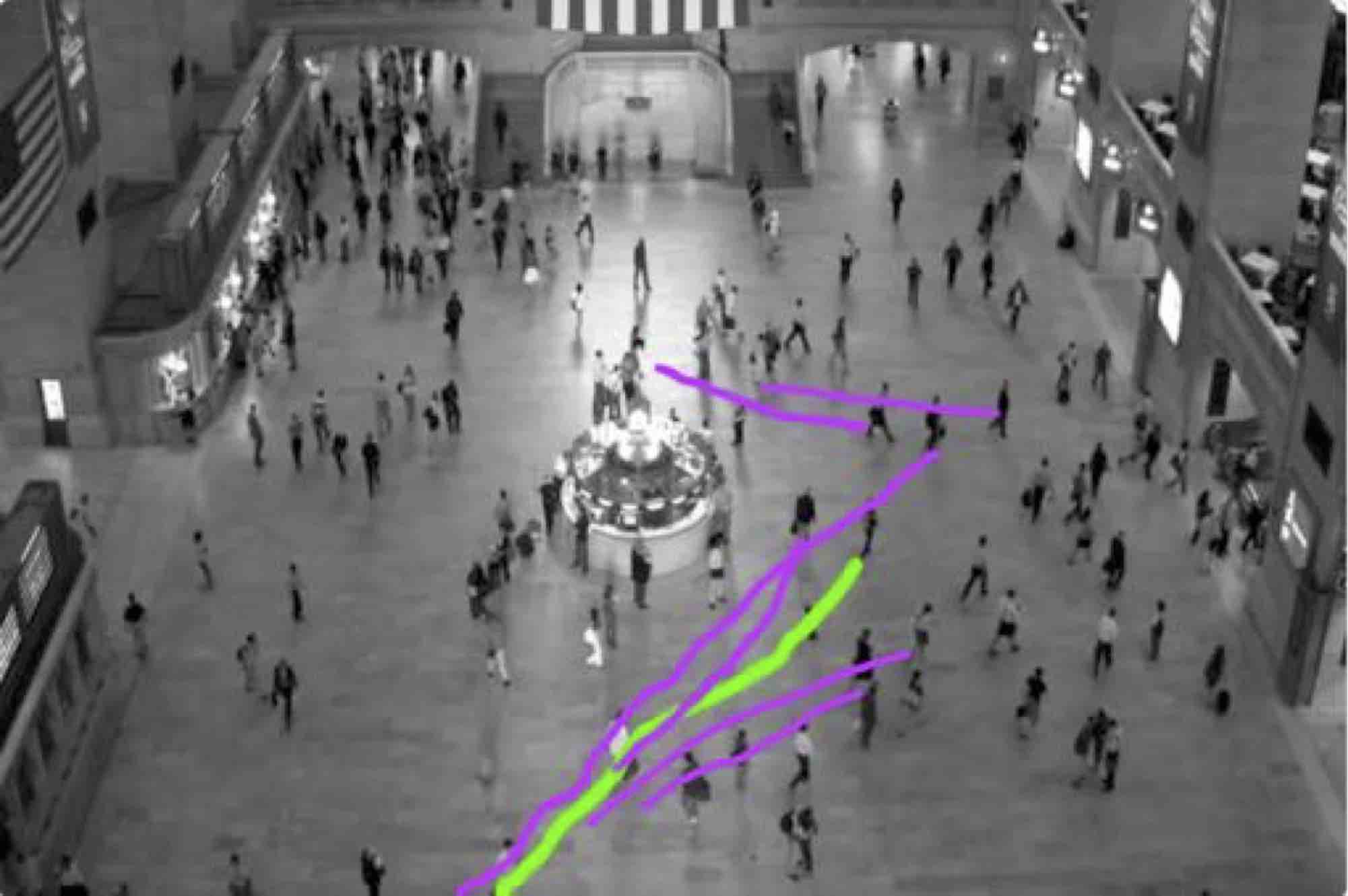}
        \caption{}
    \end{subfigure} 
    \begin{subfigure}[t]{0.23\textwidth}
        \centering
        \includegraphics[width=.95\textwidth]{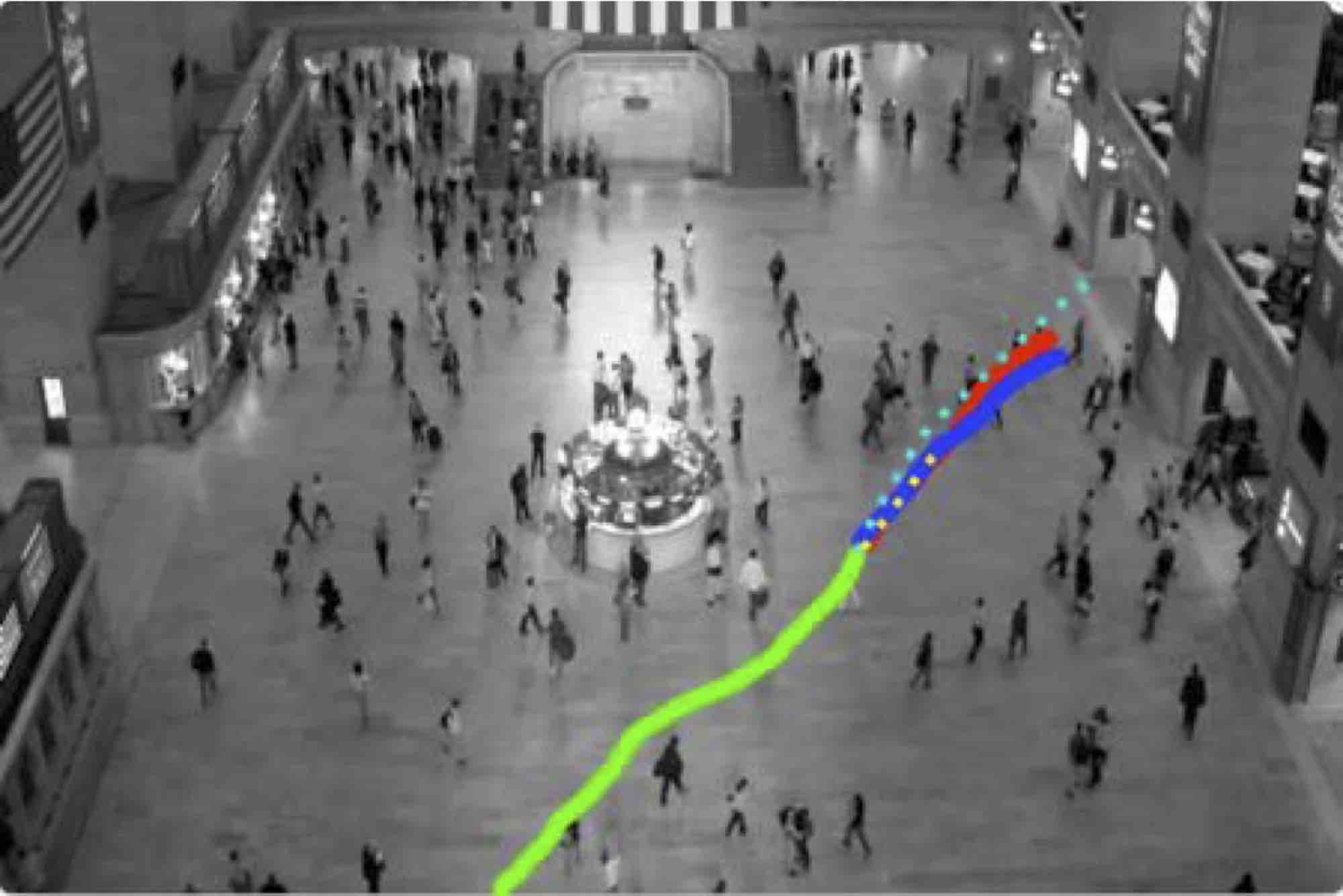}
        \caption{}
    \end{subfigure} 
    \begin{subfigure}[t]{0.23\textwidth}
        \centering
        \includegraphics[width=.95\textwidth,trim={2cm 5cm 8cm 2.80cm},clip]{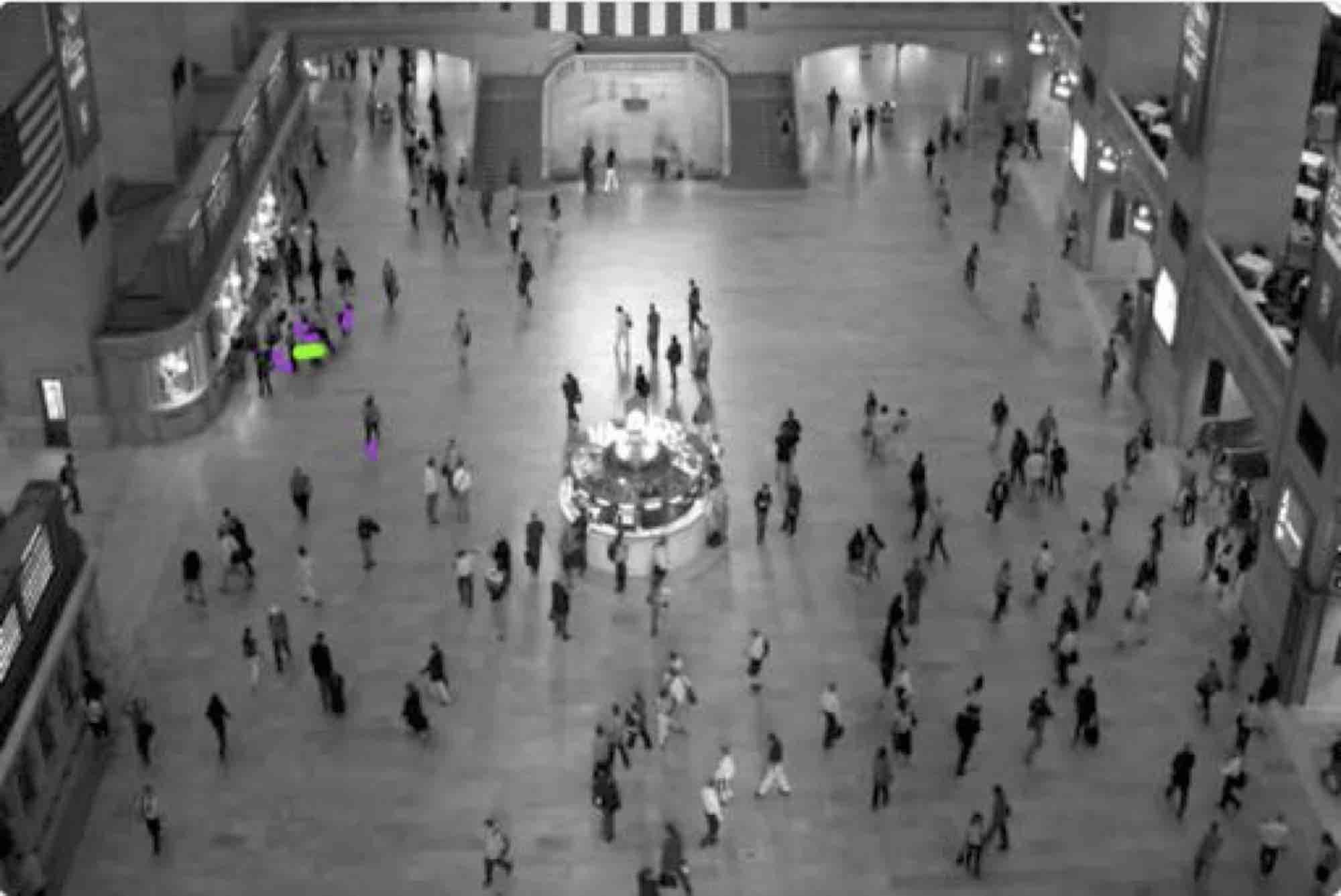} 
        \caption{}
    \end{subfigure}%
    \begin{subfigure}[t]{0.23\textwidth}
        \centering
        \includegraphics[width=.95\textwidth,trim={2cm 5cm 8cm 2.60cm},clip]{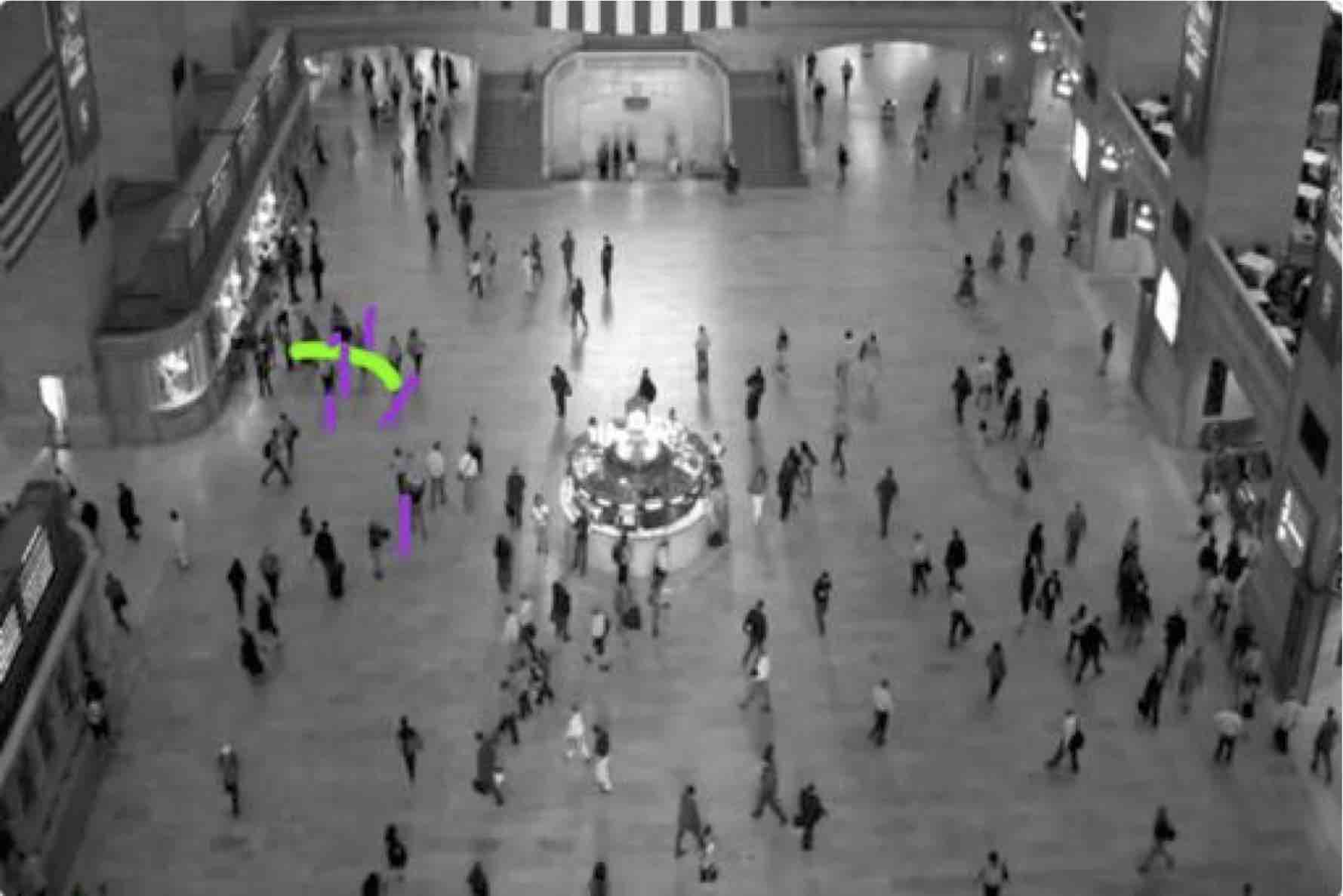} 
        \caption{}
    \end{subfigure}
     \begin{subfigure}[t]{0.23\textwidth}
        \centering
        \includegraphics[width=.95\textwidth,trim={2cm 5cm 8cm 2.80cm},clip]{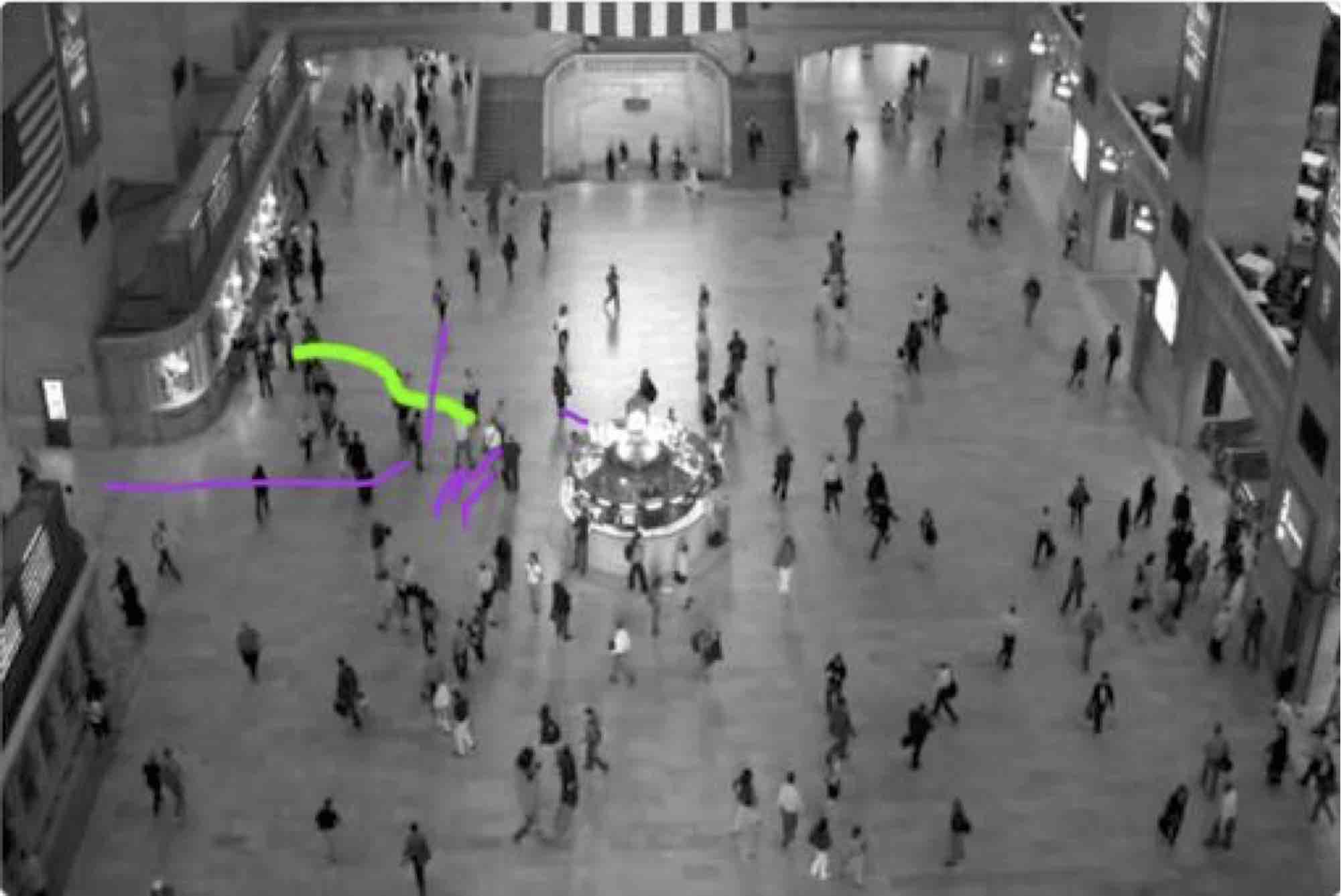}
        \caption{}
    \end{subfigure} 
        \begin{subfigure}[t]{0.23\textwidth}
        \centering
        \includegraphics[width=.95\textwidth,trim={2cm 5cm 8cm 2.60cm},clip]{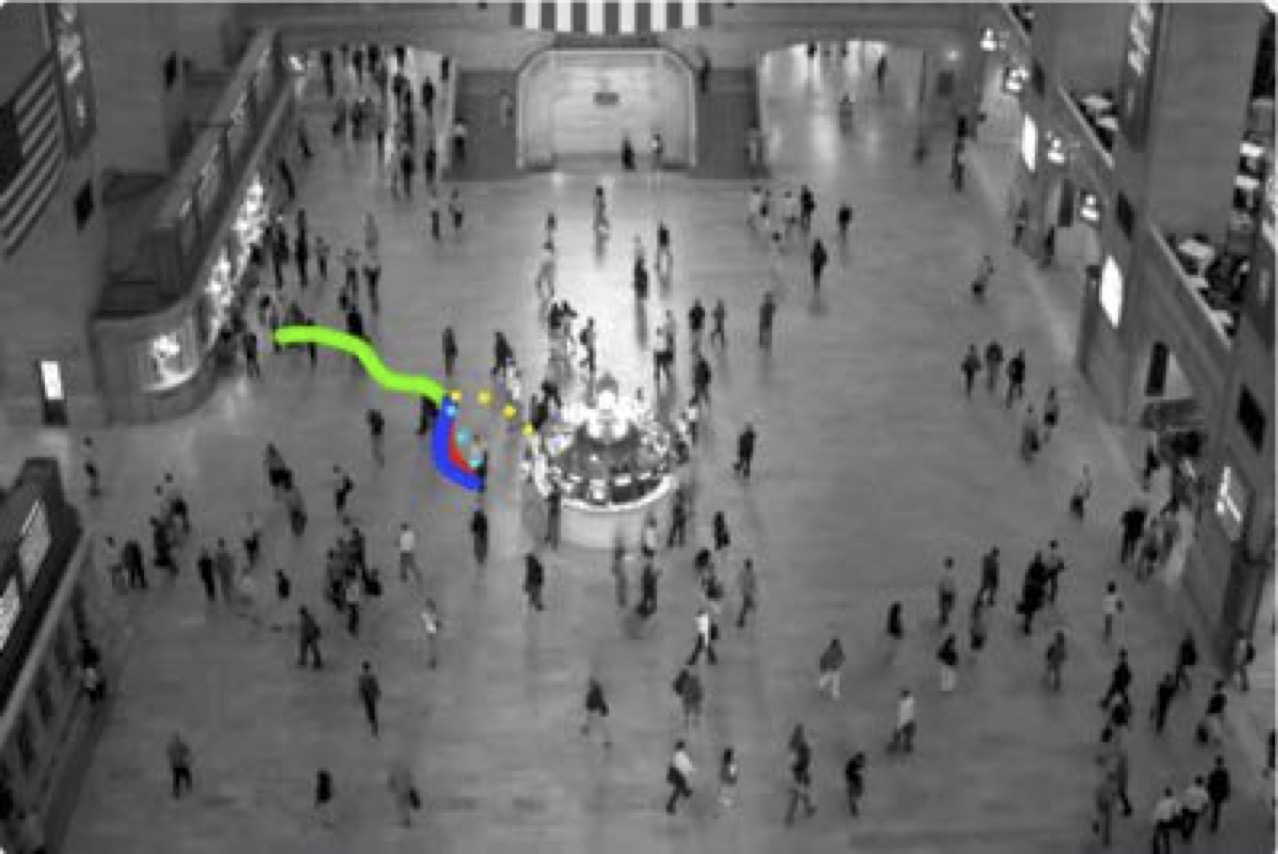}
        \caption{}
    \end{subfigure} 
    \begin{subfigure}[t]{0.23\textwidth}
        \centering
        \includegraphics[width=.95\textwidth,trim={4cm 2cm 7cm 6cm},clip]{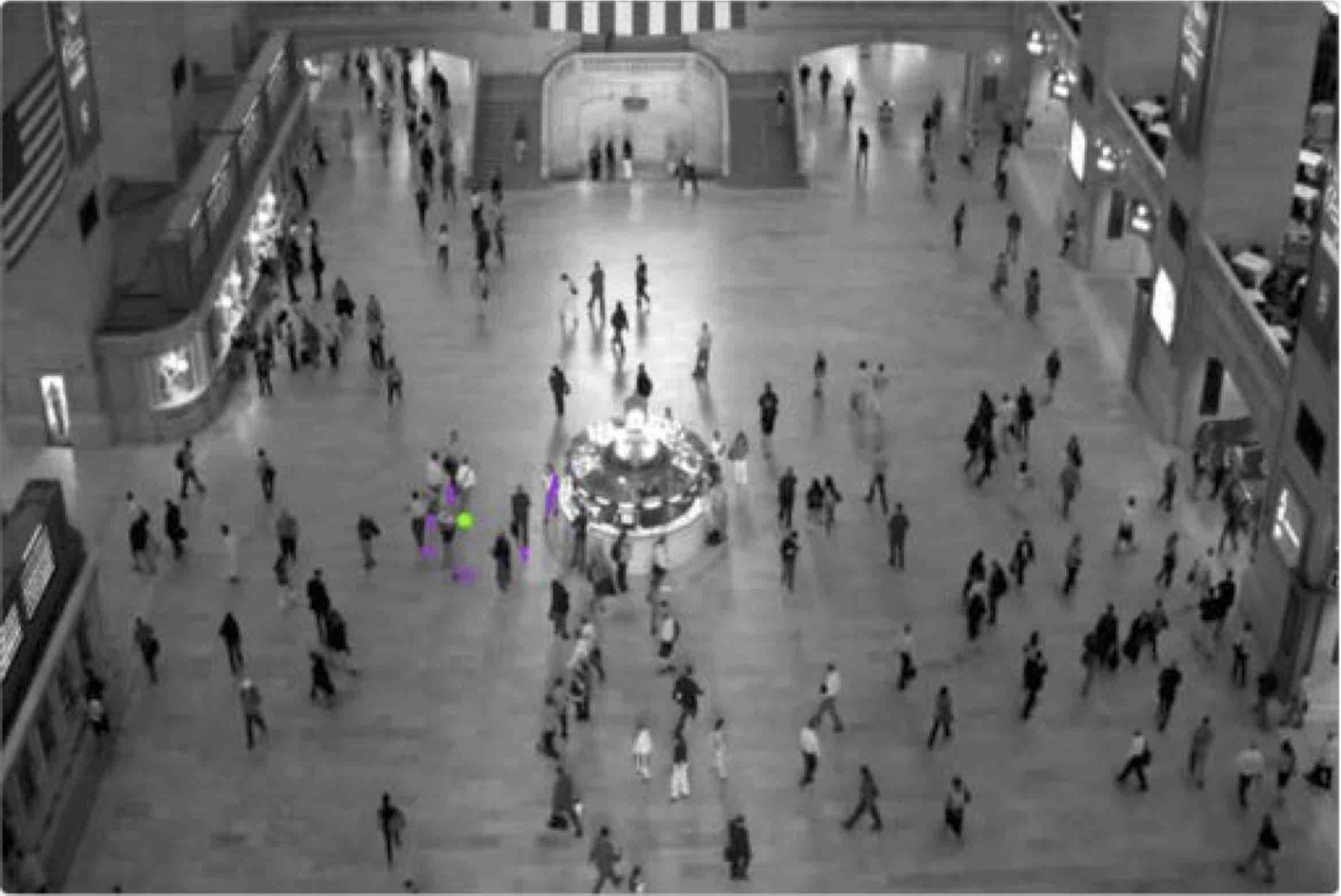} 
        \caption{}
    \end{subfigure}%
    \begin{subfigure}[t]{0.23\textwidth}
        \centering
        \includegraphics[width=.95\textwidth,trim={4cm 2cm 7cm 5.8cm},clip]{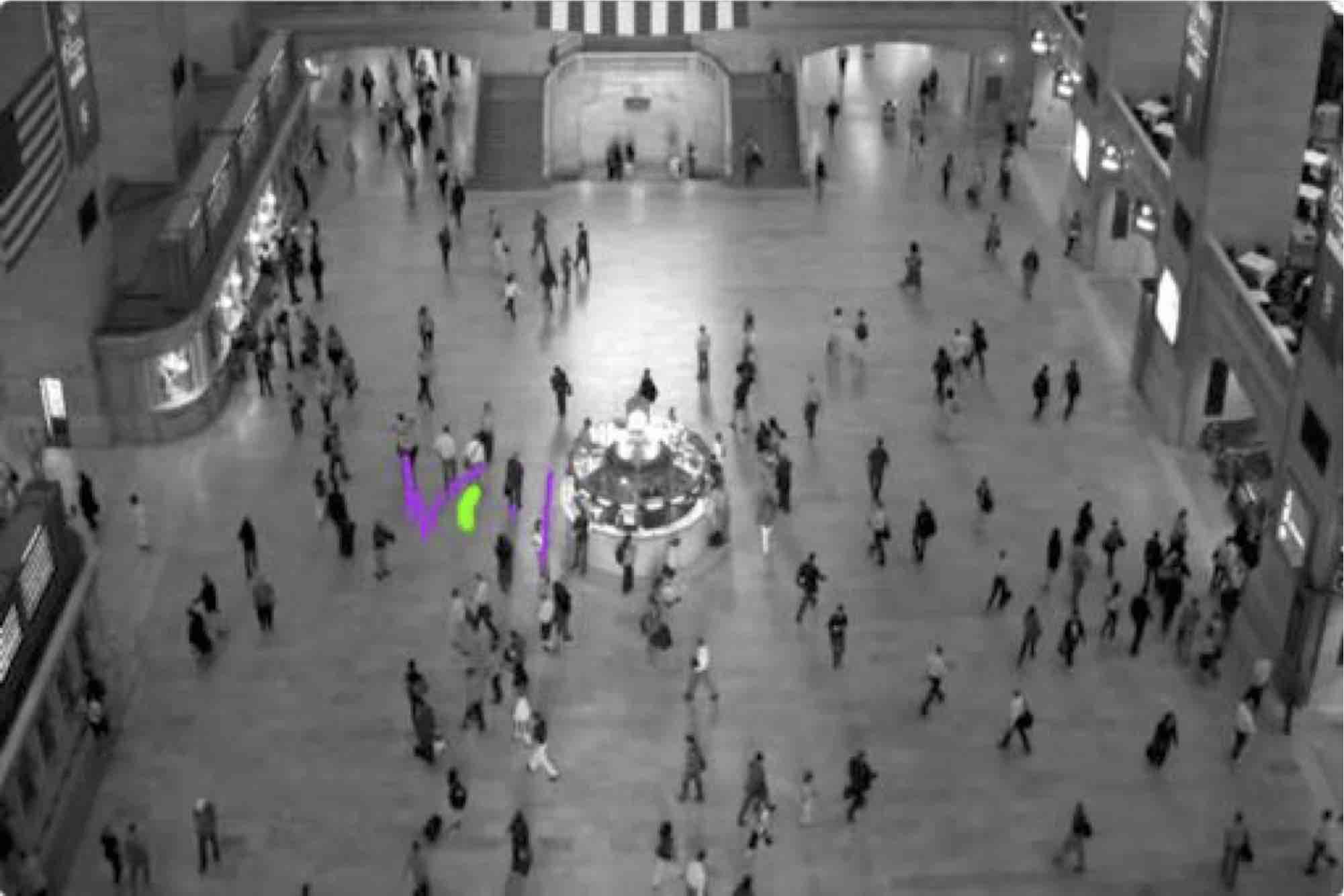} 
        \caption{}
    \end{subfigure}
     \begin{subfigure}[t]{0.23\textwidth}
        \centering
        \includegraphics[width=.95\textwidth,trim={4cm 2cm 7cm 6cm},clip]{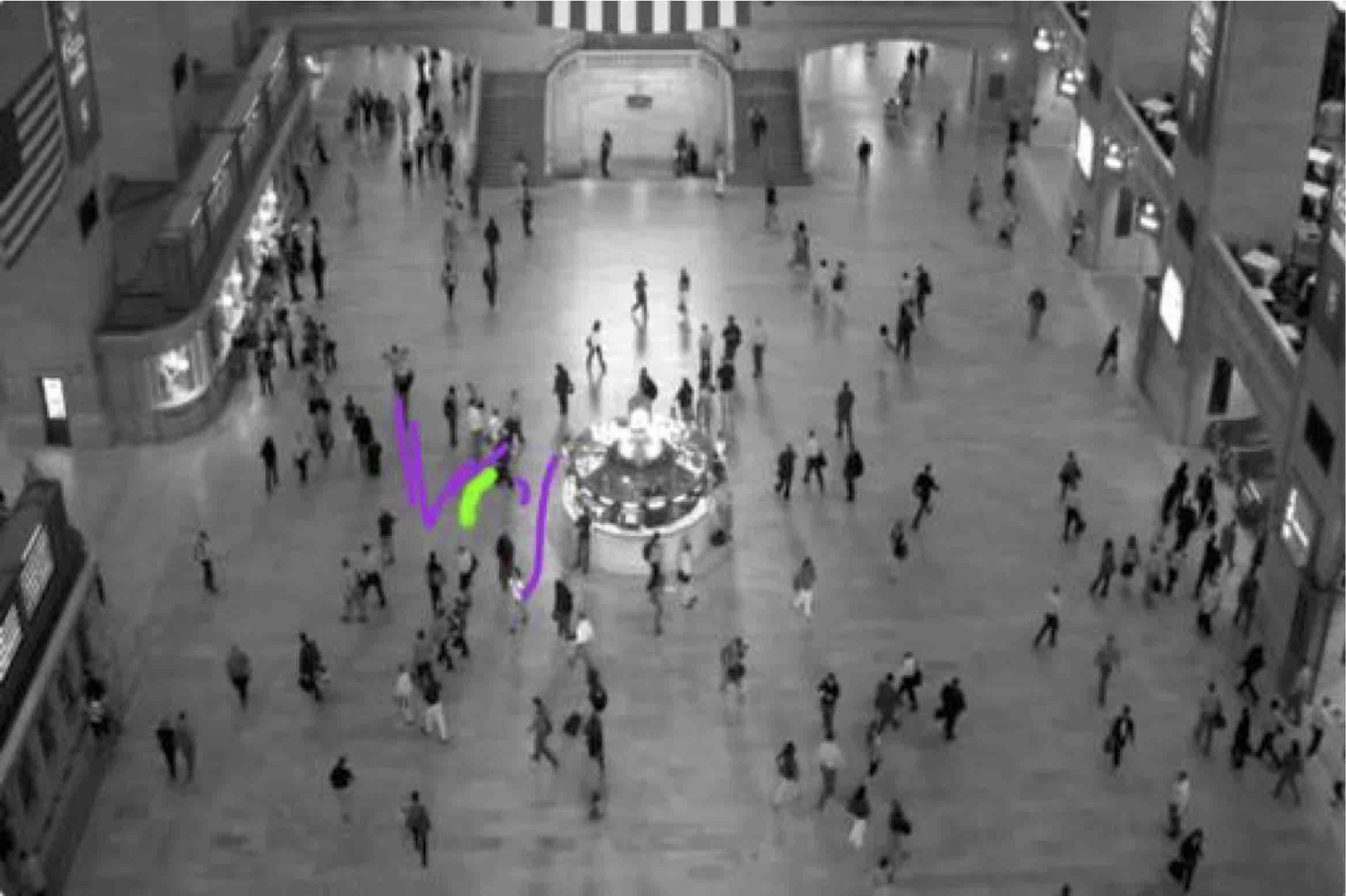}
        \caption{}
    \end{subfigure} 
         \begin{subfigure}[t]{0.23\textwidth}
        \centering
        \includegraphics[width=.95\textwidth,trim={4cm 2cm 7cm 5.8cm},clip]{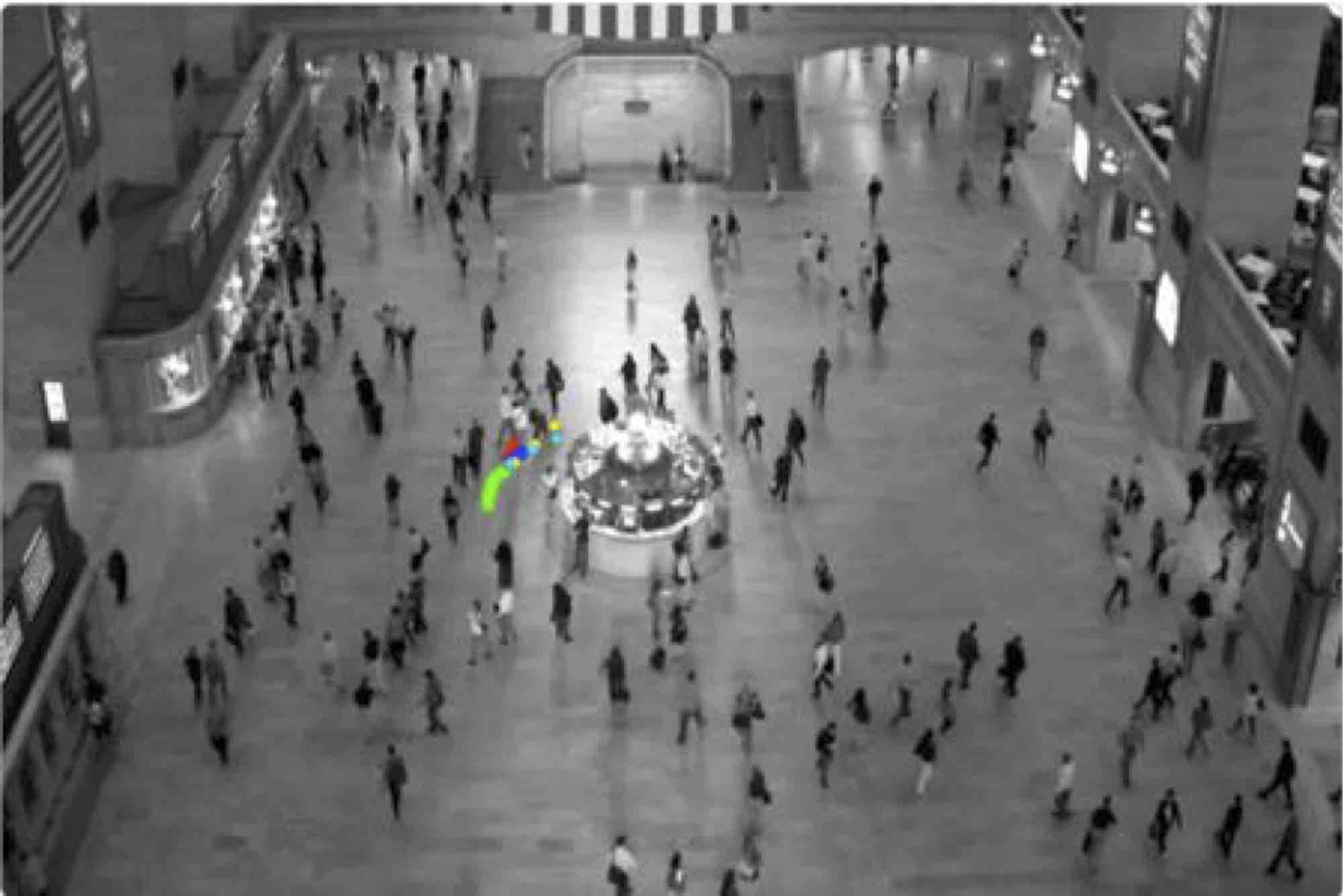}
        \caption{}
    \end{subfigure} 
    \vspace{-2mm}
    \caption{Example scenarios: Columns (left to right):  first observation of the trajectory; half way through; last observation prior to prediction; prediction from the respective models. $1^{st}$ row shows an example where the pedestrian exhibits 2 modes of motion (walking and running) within the same trajectory. $2^{nd}$  row shows an example where the previous context of the pedestrian is useful in prediction. $3^{rd}$ row shows an example where the  pedestrian is moving as part of a group. Colours: Given trajectory (in green), Ground Truth (in Blue), Neighbouring (in purple) and Predicted trajectories from $\mathbf{OUR_{cmb}}$ model (in red), from \textbf{S-LSTM} model (in yellow), from $\mathbf{OUR_{sc}}$ model (in cyan). In order to preserve visibility without occlusions we have plotted only the closest 2 neighbours in each direction.}
    \label{fig:example_scenarios}
\end{figure*}

Three example scenarios that illustrates the advantage of attending to all the hidden states within that particular context are shown in Fig. \ref{fig:example_scenarios}. The first row shows an example where the pedestrian of interest exhibits 2 modes of motion (walking and running) within the same trajectory. In the second row we have an example where the previous context of the pedestrian of interest is useful in tthe final prediction. Third row shows an example where pedestrian exhibits a group motion. The first three columns show the progression of the motion from the start of the trajectory (first column) to the point directly before the prediction is made (third column). The given trajectory of the person of interest is shown in green and the neighbouring trajectories are shown in purple. In order to preserve visibility without occlusions we have plotted only the closest 2 neighbours in each direction. Finally in the fourth column we have presented the respective predictions from $\mathbf{OUR_{cmb}}$ model (in red), from \textbf{S-LSTM} model (in yellow), from $\mathbf{OUR_{sc}}$ model (in cyan). The ground truth observations are shown in blue. \par
When considering the example shown in  Fig. \ref{fig:example_scenarios} (a)-(d) the pedestrian of interest exhibits a running behaviour when entering the scene (Fig. \ref{fig:example_scenarios} (a)). Then at half way through her trajectory she shifts her behaviour to walking (Fig. \ref{fig:example_scenarios} (b)). Therefore at the time of the prediction (Fig. \ref{fig:example_scenarios} (c)), the hidden states of the baseline model will be dominated by the walking behaviour as it is the most recent behaviour of this particular pedestrian. Therefore the predictions generated by the \textbf{S-LSTM} model are erroneous. But as we are attending to all the previous hidden states before predicting the future sequence, the multi model nature of that pedestrian's motion has been captured by the $\mathbf{OUR_{sc}}$ and  $\mathbf{OUR_{cmb}}$ models.  \par
Another example scenario where the proposed model out performs it's baseline is shown in Fig. \ref{fig:example_scenarios} (e)-(h). When deciding upon whether to give way or to cut through the pedestrian group, models  $\mathbf{OUR_{sc}}$ and  $\mathbf{OUR_{cmb}}$ outperforms the \textbf{S-LSTM} model. While attending to the historical data in \ref{fig:example_scenarios} (f) we can see the behaviour of the pedestrian of interest under a similar context. Therefore the proposed models can anticipate that the preferred behaviour of the pedestrian of interest under such context is giving way to the others. But in the \textbf{S-LSTM} model as it's attending only to the immediate preceding hidden states and those long range dependencies are not captured. \par 
In the example shown in Fig. \ref{fig:example_scenarios} (i)-(l) we illustrate the importance of considering the entire history of the neighbours. The way that the neighbours affect to a pedestrian during group motion vastly differs to the impact group motion has on a pedestrian walking alone. For example pedestrians moving as part of a group tend to walk at a similar velocity, keeping small distances between themselves, stopping or turing together; where as pedestrians walking alone try to keep a safe distance between themselves and their neighbours to avoid collisions. These notions can be quickly captured while observing the neighbourhood history. Therefore the predictions generated by both $\mathbf{OUR_{sc}}$ and  $\mathbf{OUR_{cmb}}$ models outperforms the baseline \textbf{S-LSTM} model which only considered the immediate preceding hidden state of the neighbours when generating the predictions. \par
When comparing the predictions from $\mathbf{OUR_{sc}}$ model against $\mathbf{OUR_{cmb}}$ model it is evident that the more spatial context specific predictions are generated by $\mathbf{OUR_{cmb}}$ as it has been specifically trained on the examples from that particular spatial region. Therefore it anticipates the motion of a particular pedestrian more accurately. Still in all the example scenarios $\mathbf{OUR_{sc}}$ is shown to be capable of generating acceptable predictions compared to the baseline model, showing that the proposed combined attention mechanism is capable of generating more accurate and realistic trajectories than the current state-of-the-art.

\section{Abnormal behaviour detection}
The proposed framework can be directly applied for detecting abnormal pedestrian behaviour. A naive approach would be to predict the trajectory for the period of ${T_{obs+1}}$ to ${T_{pred}}$ while observing the same trajectory over this time period and measuring the deviation between the observed and the predicted trajectories. If the deviation is greater than a threshold, then an abnormality can be said to have occurred. However due to the adaptive nature of deep neural networks, abnormal behaviours such as: i) sudden turns and changes in walking directions; and ii) trajectories with abnormal velocities; may not be classified as abnormal events. \par

We observe that the hidden states of the LSTM encoder decoder framework hold vital information which is used to model the walking behaviour of the pedestrian of interest. Hence, if his or her behaviour is abnormal then the hidden state values for that pedestrian should be distinct from those of a normal pedestrian. \par
With that intuition we randomly selected 500 trajectories from the Grand Central dataset and predicted the trajectories for those pedestrians. The trajectories were hand labeled for abnormal behaviour, considering sudden turns and changes in walking direction and abnormal velocities as the set of abnormal behaviours. The dataset consists of 445 normal trajectories and 55 abnormal trajectories. Then we extracted the encoded hidden states  ($h_{(t)}=[h_{(1)}, \ldots, h_{(T_{obs})}]$) for the given trajectory for that pedestrian and the hidden states used for decoding ($s_{(t)}=[s_{(T_{obs+1})}, \ldots, s_{(T_{pred})}]$). The resultant hidden states are passed through DBSCAN to detect outliers. With the proposed approach we detected 441 trajectories as being normal and 59 trajectories as abnormal. The resultant detections are given in Table \ref{tab:abnormal_prposed}.

\begin{table}[htpb]
\begin{center}
\begin{tabular}{l|l|c|c|c} 
\multicolumn{2}{c}{}&\multicolumn{2}{c}{Ground Truth}&\\
\cline{3-4}
\multicolumn{2}{c|}{}&Abnormal&Normal\\ 
\cline{2-4}
\multirow{2}{*}{Predicted }& Abnormal & $\textbf{47}$ & $12$\\ 
\cline{2-4}
& Normal & $8$ & $\textbf{433}$ \\ 
\cline{2-4}
\multicolumn{1}{c}{} & \multicolumn{1}{c}{Total} & \multicolumn{1}{c}{$\textbf{55}$} & \multicolumn{1}{c}{$\textbf{445}$} \\ 
\end{tabular}
\caption{Abnormal Event detection with the proposed algorithm: This approach has detected 47 out of 55 ground truth abnormal events}
\label{tab:abnormal_prposed}
\end{center}
\end{table}
Analysing the classification results, we see that false alarms are mainly due to behaviours that are erroneously detected as abnormal being uncommon in the database. Cases such as people changing direction to buy tickets, and passengers wandering in the free area are detected as abnormal due to the fact that they are not significantly present in the subset of trajectories selected for this task. 
\par
We compare this approach to the naive approach given above. It is evident that some abnormal trajectories are misclassified as normal behaviour due to its lack of deviation from the observed trajectory. See Tab. \ref{tab:abnormal_naive}

\begin{table}[htpb]
\begin{center}
\begin{tabular}{l|l|c|c|c}
\multicolumn{2}{c}{}&\multicolumn{2}{c}{Ground Truth}&\\
\cline{3-4}
\multicolumn{2}{c|}{}&Abnormal&Normal\\ 
\cline{2-4}
\multirow{2}{*}{Predicted }& Abnormal & $\textbf{29}$ & $24$ \\ 
\cline{2-4}
& Normal & $26$ & $\textbf{421}$ \\ 
\cline{2-4}
\multicolumn{1}{c}{} & \multicolumn{1}{c}{Total} & \multicolumn{1}{c}{$\textbf{55}$} & \multicolumn{1}{c}{$\textbf{445}$} \\ 
\end{tabular}
\caption{Abnormal event detection with naive approach: This approach has detected only 29 out of 55 ground truth abnormal events}
\label{tab:abnormal_naive}
\end{center}
\end{table}

\begin{figure*}[!htb]
    \centering
    \begin{subfigure}{0.3\textwidth}
        \centering
        \includegraphics[width=.95\textwidth]{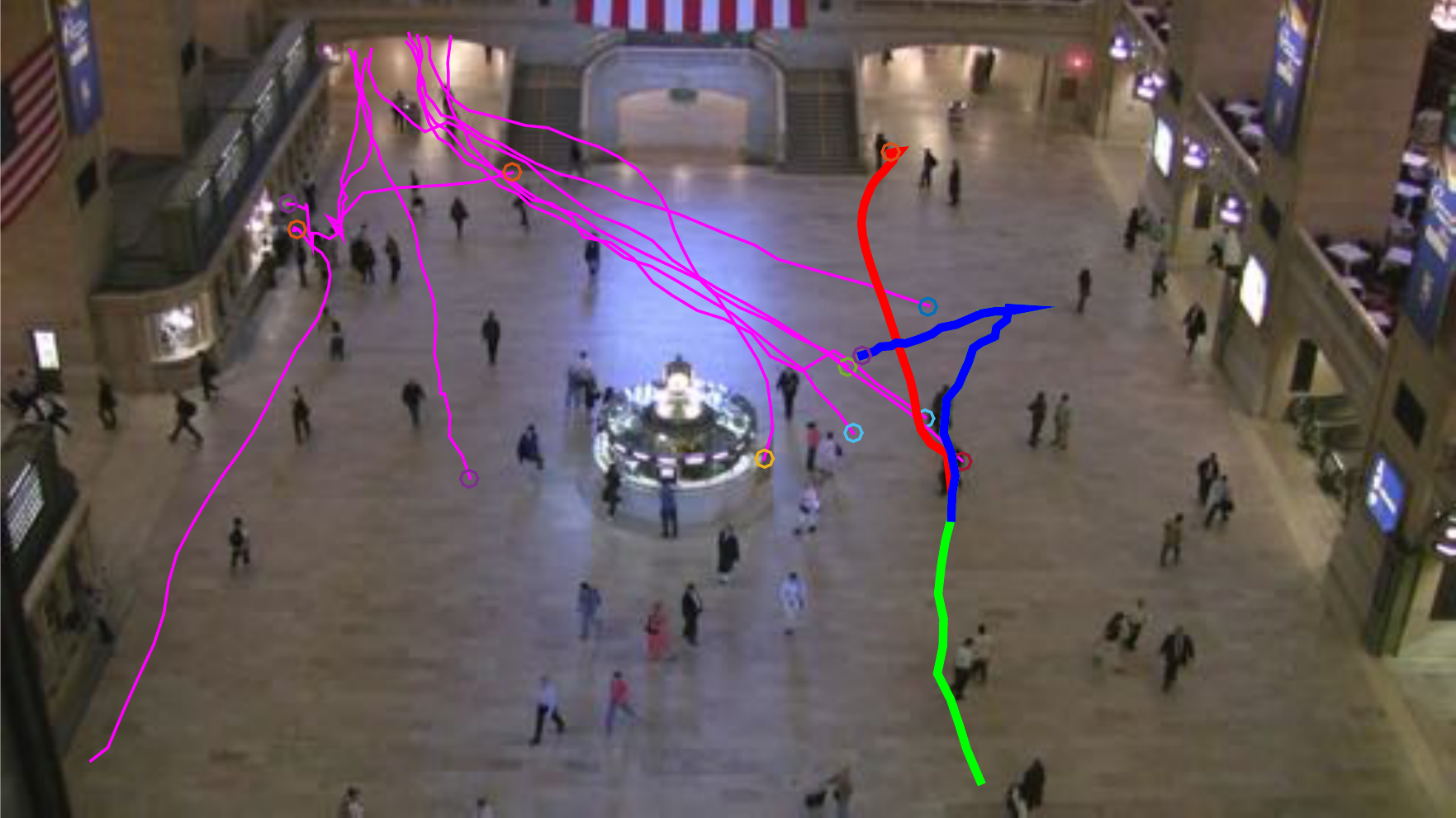}
        \caption{}
    \end{subfigure}%
    \begin{subfigure}{0.3\textwidth}
        \centering
        \includegraphics[width=.95\textwidth]{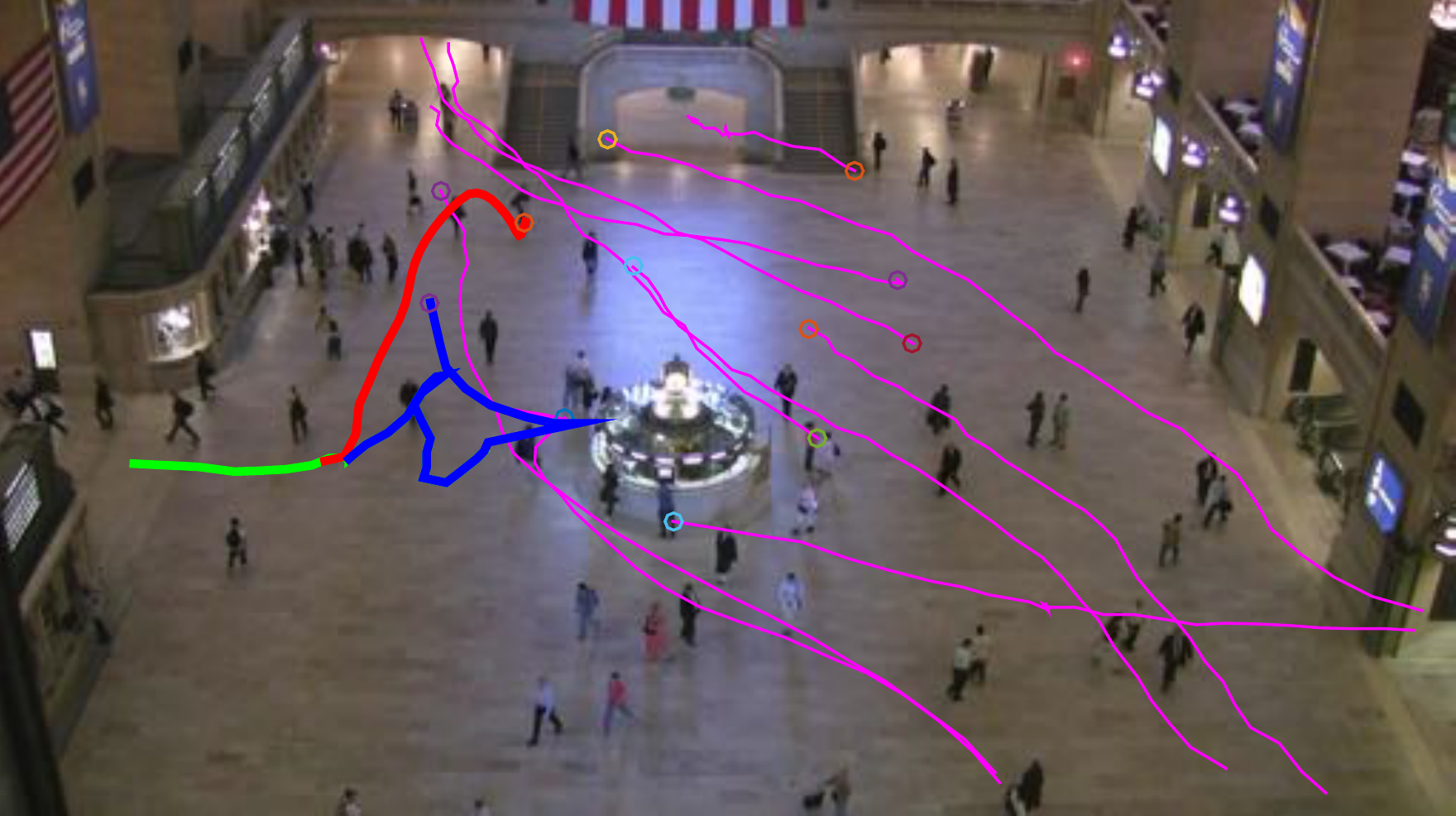}
        \caption{}
    \end{subfigure}
     \begin{subfigure}{0.3\textwidth}
        \centering
        \includegraphics[width=.95\textwidth]{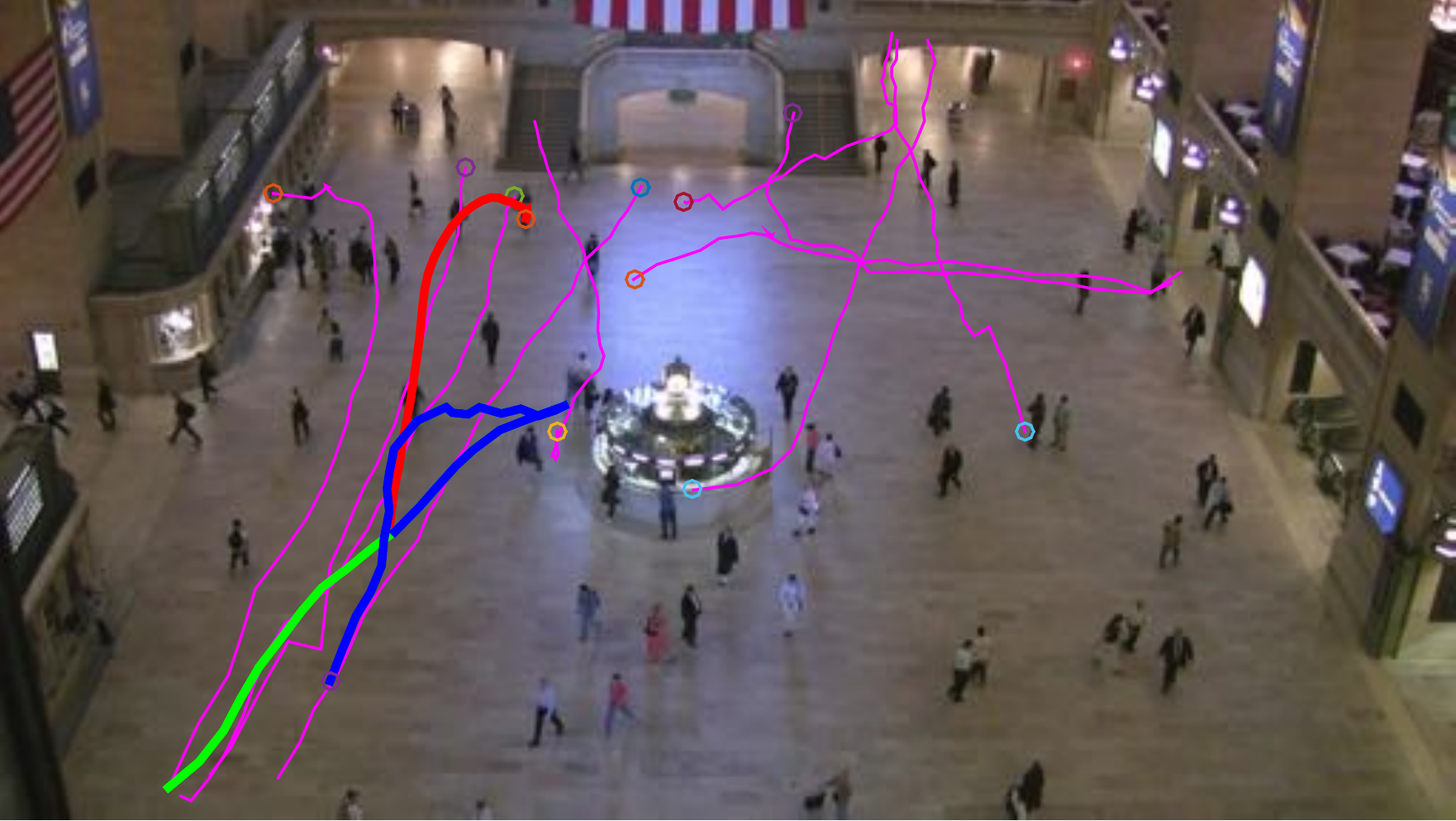}
        \caption{}
		\vspace{-1mm}
    \end{subfigure}
    \begin{subfigure}{0.3\textwidth}
        \centering
        \includegraphics[width=.95\textwidth]{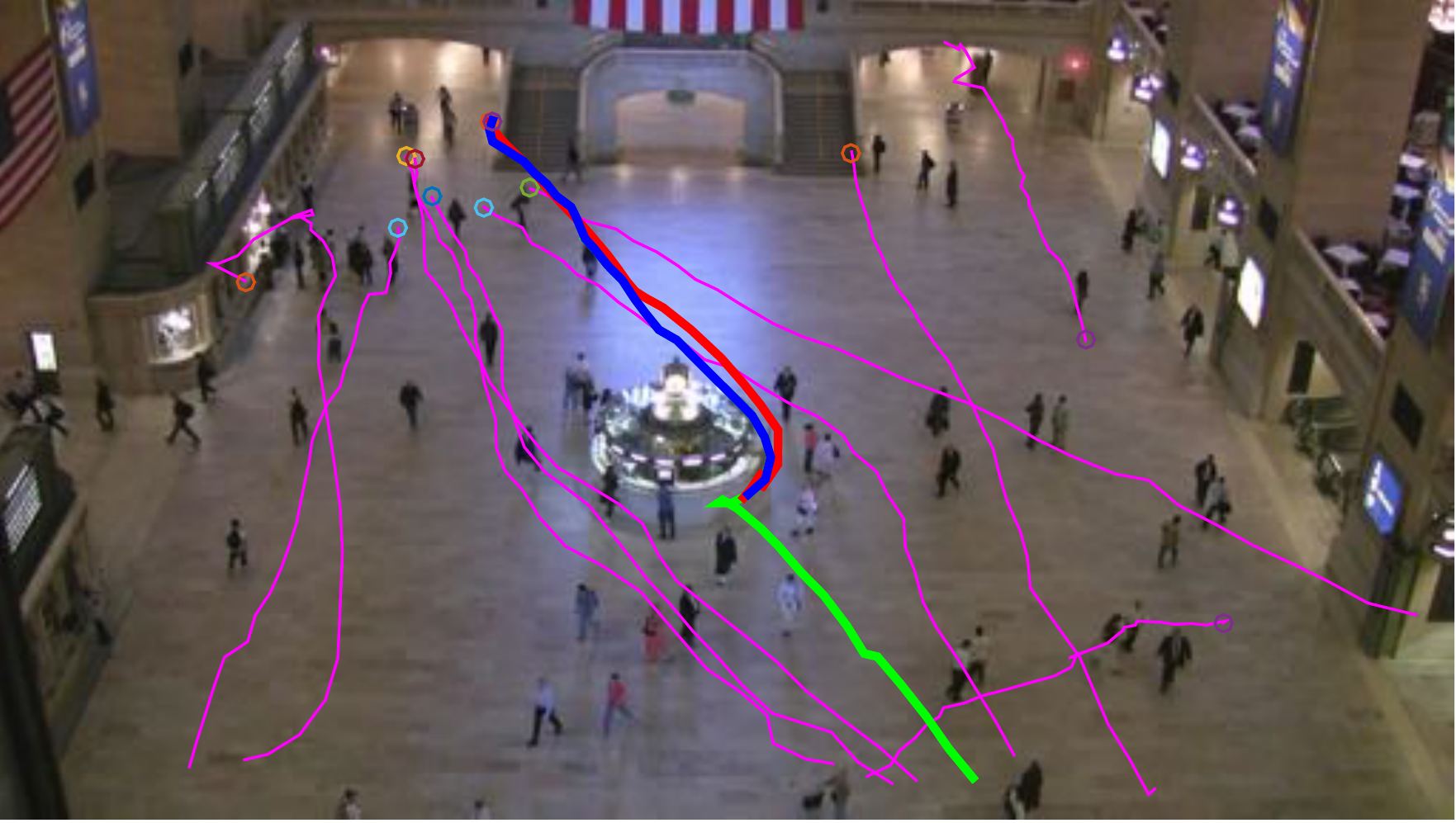}
        \caption{}
    \end{subfigure}
     \begin{subfigure}{0.3\textwidth}
        \centering
        \includegraphics[width=.95\textwidth]{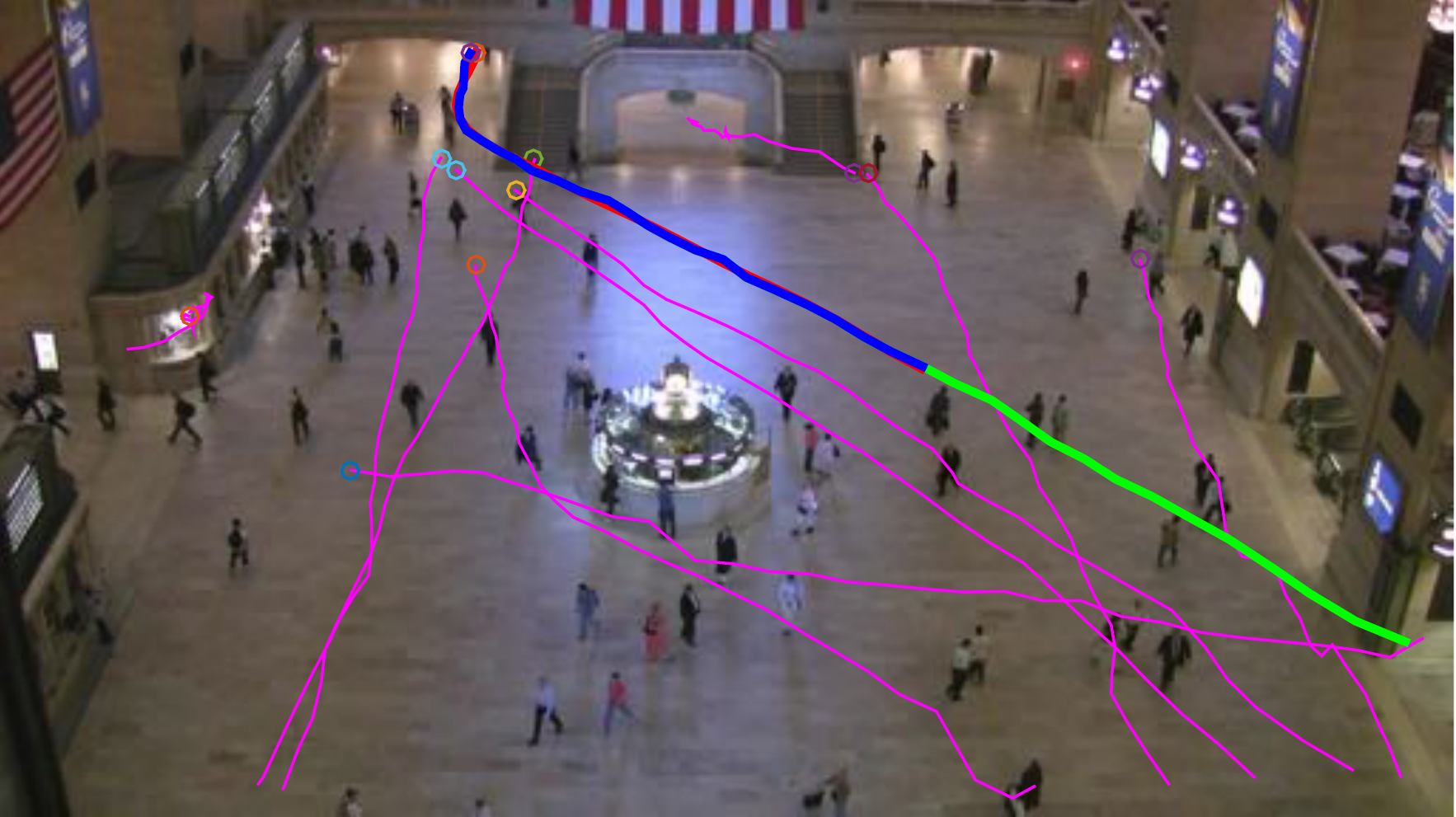}
        \caption{}
    \end{subfigure}
		\vspace{-1mm}
    \caption{Abnormal event detections:  (a)-(c) abnormal behaviour detected due to sudden change of moving direction. Abnormal behaviour due to sudden circular turn (d) and abnormal velocity in (e)}
    \label{fig:abnormal_events}
\end{figure*}
Some examples of detected abnormal events are shown in Fig. \ref{fig:abnormal_events}. The first row shows the abnormal behaviour detected due to sudden change of moving direction. Even though, in the examples shown (d) and (e), there isn't a significant deviation between the predicted path and the observed path, our abnormal event detection approach has accurately classified the event due to the sudden circular turn in the trajectory in (d) and abnormal velocity in (e).

\section{Conclusion}
In this paper we have proposed a novel neural attention based framework to model pedestrian flow in a surveillance setting. We extend the classical encoder-decoder framework in sequence to sequence modelling to incorporate both soft attention as well as hard-wired attention. This has a major positive impact when handling longer trajectories in heavily cluttered neighbourhoods. The hand-crafted hard-wired attention weights approximate the neighbour's influence and make the application of attention models pursuable for real world scenarios with large number of neighbours.  We tested our proposed model in two challenging publicly available surveillance datasets and demonstrated state-of-the-art performance. Our new neural attention framework exhibited a stronger ability to accurately predict pedestrian motion, even in the presence of multiple source and sink positions and with high crowd densities observed. Furthermore, we have shown how the proposed approach can support abnormal event detection through hidden state clustering. This approach is able to accurately detect events in challenging situations, without handcrafting the features.
Apart from direct applications such as abnormal behaviour detection, improving passenger flow in transport environments, this framework can be extended to any application domain where modelling multiple co-occurring trajectories is necessary. Some potential areas include modelling aircraft movements, ship trajectories and vehicle traffic.   

\section*{References}
\bibliography{mydatabase}

\end{document}